\documentclass{article}

\pdfoutput=1

\PassOptionsToPackage{numbers, compress}{natbib}
\usepackage[final]{nips_2016}

\usepackage[utf8]{inputenc} 
\usepackage[T1]{fontenc}    
\usepackage{hyperref}       
\usepackage{url}            
\usepackage{booktabs}       
\usepackage{amsfonts}       
\usepackage{nicefrac}       
\usepackage{microtype}      

\usepackage{floatrow}       

\usepackage{amsmath}
\usepackage{amsthm}
\usepackage{amssymb}
\usepackage{enumitem}
\usepackage{mathtools}
\usepackage{times}
\usepackage{graphicx} 
\usepackage{subcaption}
\usepackage{stackengine}

\usepackage{siunitx}
\usepackage{lscape}

\usepackage{algorithm}
\usepackage{algorithmic}

\usepackage{hyperref}

\newcommand{\K}{\!\{\!K\!\}}
\newcommand{\KL}{\!\{\!K_L\!\}}
\newcommand{\W}{\scriptscriptstyle W}
\newfloatcommand{capbtabbox}{table}[][\FBwidth]

\newtheorem{definition}{Definition}

\title{Scaling Memory-Augmented Neural Networks with Sparse Reads and Writes}

\author{
Jack W Rae\footnotemark[1] \\
\texttt{jwrae}
\And
Jonathan J Hunt\footnotemark[1] \\
\texttt{jjhunt}
\And
Tim Harley\\
\texttt{tharley}
\And
Ivo Danihelka \\
\texttt{danihelka}
\And
Andrew Senior \\
\texttt{andrewsenior}
\And
Greg Wayne \\
\texttt{gregwayne}
\And
Alex Graves \\
\texttt{gravesa}
\And
Timothy P Lillicrap \\
\texttt{countzero}
\\
\\
\hspace{-8.3cm}
Google DeepMind \\
\hspace{-8.3cm}
\texttt{@google.com}
}

\begin{document}

\maketitle

\renewcommand{\thefootnote}{\fnsymbol{footnote}}
\footnotetext[1]{These authors contributed equally.}


\begin{abstract}
Neural networks augmented with external memory have the ability to learn algorithmic solutions to complex tasks. These models appear promising for applications such as language modeling and machine translation. However, they scale poorly in both space and time as the amount of memory grows --- limiting their applicability to real-world domains. Here, we present an end-to-end differentiable memory access scheme, which we call Sparse Access Memory (SAM), that retains the representational power of the original approaches whilst training efficiently with very large memories. We show that SAM achieves asymptotic lower bounds in space and time complexity, and find that an implementation runs $1,\!000\times$ faster and with $3,\!000\times$ less physical memory than non-sparse models. SAM learns with comparable data efficiency to existing models on a range of synthetic tasks and one-shot Omniglot character recognition, and can scale to tasks requiring $100,\!000$s of time steps and memories. As well, we show how our approach can be adapted for models that maintain temporal associations between memories, as with the recently introduced Differentiable Neural Computer.


\end{abstract}

\section{Introduction}

Recurrent neural networks, such as the Long Short-Term Memory (LSTM) \cite{hochreiter1997long}, have proven to be powerful sequence learning models \cite{graves2013speech, sutskever2014}. However, one limitation of the LSTM architecture is that the number of parameters grows proportionally to the square of the size of the memory, making them unsuitable for problems requiring large amounts of long-term memory. Recent approaches, such as Neural Turing Machines (NTMs) \cite{graves2014neural} and Memory Networks \cite{weston2014memory}, have addressed this issue by decoupling the memory capacity from the number of model parameters. We refer to this class of models as memory augmented neural networks (MANNs). External memory allows MANNs to learn algorithmic solutions to problems that have eluded the capabilities of traditional LSTMs, and to generalize to longer sequence lengths. Nonetheless, MANNs have had limited success in real world application.

A significant difficulty in training these models results from their smooth read and write operations, which incur linear computational overhead on the number of memories stored per time step of training. Even worse, they require duplication of the entire memory at each time step to perform backpropagation through time (BPTT). To deal with sufficiently complex problems, such as processing a book, or Wikipedia, this overhead becomes prohibitive. For example, to store $64$ memories, a straightforward implementation of the NTM trained over a sequence of length $100$ consumes $\approx \SI{30}{MiB}$ physical memory; to store $64,\!000$ memories the overhead exceeds $\SI{29}{GiB}$ (see Figure \ref{fig:perf_benchmarks}).

In this paper, we present a MANN named SAM (sparse access memory). By thresholding memory modifications to a sparse subset, and using efficient data structures for content-based read operations, our model is optimal in space and time with respect to memory size, while retaining end-to-end gradient based optimization. To test whether the model is able to learn with this sparse approximation, we examined its performance on a selection of synthetic and natural tasks: algorithmic tasks from the NTM work \cite{graves2014neural}, Babi reasoning tasks used with Memory Networks \cite{sukhbaatar2015end} and Omniglot one-shot classification \cite{santoro2016, lake2015human}. We also tested several of these tasks scaled to longer sequences via curriculum learning.  For large external memories we observed improvements in empirical run-time and memory overhead by up to three orders magnitude over vanilla NTMs, while maintaining near-identical data efficiency and performance.

Further, in Supplementary \ref{sec:sdnc} we demonstrate the generality of our approach by describing how to construct a sparse version of the recently published Differentiable Neural Computer   \cite{graves2016dnc}. This Sparse Differentiable Neural Computer (SDNC) is over $400 \times$ faster than the canonical dense variant for a memory size of $2,\!000$ slots, and achieves the best reported result in the Babi tasks without supervising the memory access.

\section{Background}

\subsection{Attention and content-based addressing}

An external memory $\mathbf{M} \in \mathbb{R}^{N \times M}$ is a collection of $N$ real-valued vectors, or \textit{words}, of fixed size $M$. A soft \textit{read} operation is defined to be a weighted average over memory words,
\begin{equation}
\label{eq:soft_read}
r = \sum_{i=1}^{N} w(i) \mathbf{M}(i) \, ,
\end{equation}
where $w \in \mathbb{R}^{N}$ is a vector of weights with non-negative entries that sum to one. Attending to memory is formalized as the problem of computing $w$. A \textit{content addressable memory}, proposed in \cite{graves2014neural,weston2014memory,bahdanau2014neural,sukhbaatar2015end}, is an external memory with an addressing scheme which selects $w$ based upon the similarity of memory words to a given query $q$. Specifically, for the $i$th read weight $w(i)$ we define,
\begin{equation}
\label{eq:content_based_read}
w(i) = \frac{f\left(d(q, \mathbf{M}(i))\right)}{\sum_{j=1}^N f\left(d(q, \mathbf{M}(j)\right)}, 
\end{equation}
where $d$ is a similarity measure, typically Euclidean distance or cosine similarity, and $f$ is a differentiable monotonic transformation, typically a softmax. We can think of this as an instance of kernel smoothing where the network learns to query relevant points $q$.
Because the read operation (\ref{eq:soft_read}) and content-based addressing scheme (\ref{eq:content_based_read}) are smooth, we can place them within a neural network, and train the full model using backpropagation.



\subsection{Memory Networks}

One recent architecture, Memory Networks, make use of a content addressable memory that is accessed via a series of read operations \cite{weston2014memory,sukhbaatar2015end} and has been successfully applied to a number of question answering tasks \cite{weston2015towards,hill2015goldilocks}. In these tasks, the memory is pre-loaded using a learned embedding of the provided context, such as a paragraph of text, and then the controller, given an embedding of the question, repeatedly queries the memory by content-based reads to determine an answer.



\subsection{Neural Turing Machine}

The Neural Turing Machine is a recurrent neural network equipped with a content-addressable memory, similar to Memory Networks, but with the additional capability to write to memory over time. The memory is accessed by a controller network, typically  an LSTM, and the full model is differentiable --- allowing it to be trained via BPTT.






%
A \textit{write} to memory,
\begin{equation}
\label{eq:soft_write}
\mathbf{M_t} \leftarrow \mathbf{(1 - \mathbf{R}_t) \odot M_{t - 1}}  + \mathbf{A_t} \; ,
\end{equation}
consists of a copy of the memory from the previous time step $\mathbf{M}_{t - 1}$ decayed by the erase matrix $\mathbf{R}_t$ indicating obsolete or inaccurate content, and an addition of new or updated information $\mathbf{A}_t$. The erase matrix $\mathbf{R_t} = w^{\W}_t e_t^T$ is constructed as the outer product between a set of write weights $w^{\W}_t \in [0, 1]^N$ and erase vector $e_t \in [0, 1]^M$. The add matrix $\mathbf{A}_T = w^{\W}_t a_t^T$ is the outer product between the write weights and a new \textit{write word} $a_t \in \mathbb{R}^{M}$, which the controller outputs.

\section{Architecture}
\label{sec:architecture}

This paper introduces \textit{Sparse Access Memory (SAM)}, a new neural memory architecture with two innovations. Most importantly, all writes to and reads from external memory are constrained to a sparse subset of the memory words, providing similar functionality as the NTM, while allowing computational and memory efficient operation. Secondly, we introduce a sparse memory management scheme that tracks memory usage and finds unused blocks of memory for recording new information.

For a memory containing $N$ words, SAM executes a forward, backward step in $\Theta(\log N)$ time, initializes in $\Theta(N)$ space, and consumes $\Theta(1)$ space per time step. Under some reasonable assumptions, SAM is asymptotically optimal in time and space complexity (Supplementary \ref{sec:space_time_suppl}).







%

\subsection{Read}
The sparse read operation is defined to be a weighted average over a selection of words in memory:

\begin{equation}
\label{eq:sparse_read_forward}
\tilde r_t = \sum_{i = 1}^K   \tilde w_t^R(s_i) \mathbf{M}_t (s_i),
\end{equation}

where $ \tilde w_t^R \in \mathbb{R}^{N}$ contains $K$ number of non-zero entries with indices $s_1, s_2, \ldots, s_K$; $K$ is a small constant, independent of $N$, typically $K=4$ or $K=8$. We will refer to sparse analogues of weight vectors $w$ as $\tilde w$, and when discussing operations that are used in both the sparse and dense versions of our model use $w$.

We wish to construct $\tilde w^R_t$ such that $\tilde r_t \approx r_t$. For content-based reads where $w^R_t$ is defined by (\ref{eq:content_based_read}), an effective approach is to keep the $K$ largest non-zero entries and set the remaining entries to zero. We can compute $\tilde w^R_t$ naively in $\mathcal{O}(N)$ time by calculating $w^R_t$ and keeping the $K$ largest values.
However, linear-time operation can be avoided. Since the $K$ largest values in $w^R_t$ correspond to the $K$ closest points to our query $q_t$, we can use an approximate nearest neighbor data-structure, described in Section \ref{sec:ann}, to calculate $\tilde w^R_t$ in $\mathcal{O}(\log N)$ time.

Sparse read can be considered a special case of the matrix-vector product defined in (\ref{eq:soft_read}), with two key distinctions. The first is that we pass gradients for only  a constant $K$ number of rows of memory per time step, versus $N$, which results in a negligible fraction of non-zero error gradient per timestep when the memory is large. The second distinction is in implementation: by using an efficient sparse matrix format such as Compressed Sparse Rows (CSR), we can compute (\ref{eq:sparse_read_forward}) and its gradients in constant time and space (see Supplementary \ref{sec:space_time_suppl}).

\subsection{Write}

The write operation is SAM is an instance of (\ref{eq:soft_write}) where the write weights $\tilde w_t^{\W}$ are constrained to contain a constant number of non-zero entries. This is done by a simple scheme where the controller writes either to previously read locations, in order to update contextually relevant memories, or the \textit{least recently accessed} location, in order to overwrite stale or unused memory slots with fresh content.

The introduction of sparsity could be achieved via other write schemes. For example, we could use a sparse content-based write scheme, where the controller chooses a query vector $q_t^{\W}$ and applies writes to similar words in memory. This would allow for direct memory updates, but would create problems when the memory is empty (and shift further complexity to the controller). We decided upon the previously read / least recently accessed addressing scheme for simplicity and flexibility.

The write weights are defined as
\begin{equation}
    \label{eq:lru_write_weights}
    w^{\W}_t = \alpha_t \, \left( \gamma_t \, w^R_{t-1} + (1 - \gamma_t) \, \mathbb{I}^U_t \right) \, ,
\end{equation}

where the controller outputs the interpolation gate parameter $ \gamma_t $ and the write gate parameter $ \alpha_t$. The write to the previously read locations $w^R_{t-1}$ is purely additive, while the least recently accessed word $\; \mathbb{I}^U_t$ is set to zero before being written to. When the read operation is sparse ($w^R_{t-1}$ has $K$ non-zero entries), it follows the write operation is also sparse.

We define $\mathbb{I}^U_t$ to be an indicator over words in memory, with a value of $1$ when the word minimizes a usage measure $U_t$
\begin{equation}
    \mathbb{I}^U_t(i) = \left \{
                \begin{array}{ll}
                    1 \;\; \hbox{ if } \; U_t(i) = \! \displaystyle{\min_{j = 1, \ldots , N}} U_t(j) \\
                    0 \hbox{ otherwise.}
                \end{array} \right .
\end{equation}
If there are several words that minimize $U_t$ then we choose arbitrarily between them. We tried two definitions of $U_t$. The first definition is a time-discounted sum of write weights $\; U^{(1)}_T(i) = \sum_{t = 0}^{T} \lambda^{T-t} \, (w^{\W}_t(i) + w^R_t(i)) $ where $\lambda$ is the discount factor.
This usage definition is incorporated within \textit{Dense Access Memory} (DAM), a dense-approximation to SAM that is used for experimental comparison in Section \ref{sec:results}.

The second usage definition, used by SAM, is simply the number of time-steps since a non-negligible memory access: $\; U^{(2)}_T(i) = T - \max \, \{\, t: w^{\W}_t(i) + w^R_t(i) > \delta\} \;$.
Here, $\delta$ is a tuning parameter that we typically choose to be $0.005$. We maintain this usage statistic in constant time using a custom data-structure (described in Supplementary \ref{sec:space_time_suppl}). Finally we also use the least recently accessed word to calculate the erase matrix. $\mathbf{R}_t = \mathbb{I}^U_t \mathbf{1}^T$ is defined to be the expansion of this usage indicator where $\mathbf{1}$ is a vector of ones. The total cost of the write is constant in time and space for both the forwards and backwards pass, which improves on the linear space and time dense write (see Supplementary \ref{sec:space_time_suppl}).




\subsection{Controller}

We use a one layer LSTM for the controller throughout.  At each time step, the LSTM receives a concatenation of the external input, $x_t$, the word, $r_{t-1}$ read in the previous time step.  The LSTM then produces a vector, $p_t = (q_t, a_t, \alpha_t, \gamma_t)$, of read and write parameters for memory access via a linear layer.  The word read from memory for the current time step, $r_t$, is then concatenated with the output of the LSTM, and this vector is fed through a linear layer to form the final output, $y_t$. The full control flow is illustrated in Supplementary Figure \ref{fig:controlflow}.

\subsection{Efficient backpropagation through time}
\label{sec:efficient_bptt}

We have already demonstrated how the forward operations in SAM can be efficiently computed in $\mathcal{O}(T\log N)$ time. However, when considering space complexity of MANNs, there remains a dependence on $\mathbf{M_t}$ for the computation of the derivatives at the corresponding time step. A naive implementation 
requires the state of the memory to be cached at each time step, incurring a space overhead of $\mathcal{O}(NT)$, which severely limits memory size and sequence length.

Fortunately, this can be remedied. Since there are only $\mathcal{O}(1)$ words that are written at each time step, we instead track the sparse modifications made to the memory at each timestep, apply them in-place to compute $\mathbf{M_t}$ in $\mathcal{O}(1)$ time and $\mathcal{O}(T)$ space. During the backward pass, we can restore the state of $M_{t}$ from $M_{t+1}$ in $\mathcal{O}(1)$ time by reverting the sparse modifications applied at time step $t$. As such the memory is actually rolled back to previous states during backpropagation (Supplementary Figure \ref{fig:bptt}).

At the end of the backward pass, the memory ends rolled back to the start state. If required, such as when using truncating BPTT, the final memory state can be restored by making a copy of $\mathbf{M_T}$ prior to calling backwards in $\mathcal{O}(N)$ time, or by re-applying the $T$ sparse updates in $\mathcal{O}(T)$ time.





\subsection{Approximate nearest neighbors}
\label{sec:ann}

When querying the memory, we can use an approximate nearest neighbor index (ANN) to search over the external memory for the $K$ nearest words.
Where a linear KNN search inspects every element in memory (taking $\mathcal{O}(N)$ time), an ANN index maintains a structure over the dataset to allow for fast inspection of nearby points in $\mathcal{O}(\log N)$ time.

In our case, the memory is still a dense tensor that the network directly operates on; however the ANN is a structured view of its contents. Both the memory and the ANN index are passed through the network and kept in sync during writes. However there are no gradients with respect to the ANN as its function is fixed.

We considered two types of ANN indexes: FLANN's randomized k-d tree implementation \cite{flann_pami_2014} that arranges the datapoints in an ensemble of structured (randomized k-d) trees to search for nearby points via comparison-based search, and one that uses locality sensitive hash (LSH) functions that map points into buckets with distance-preserving guarantees. We used randomized k-d trees for small word sizes and LSHs for large word sizes.
For both ANN implementations, there is an $\mathcal{O}(\log N)$ cost for insertion, deletion and query. We also rebuild the ANN from scratch every $N$ insertions to ensure it does not become imbalanced.

\section{Results}

\label{sec:results}


\subsection{Speed and memory benchmarks}

\begin{figure*}[h]
    \centering
    \begin{subfigure}{0.47\textwidth}
    \includegraphics[height=3.4cm]{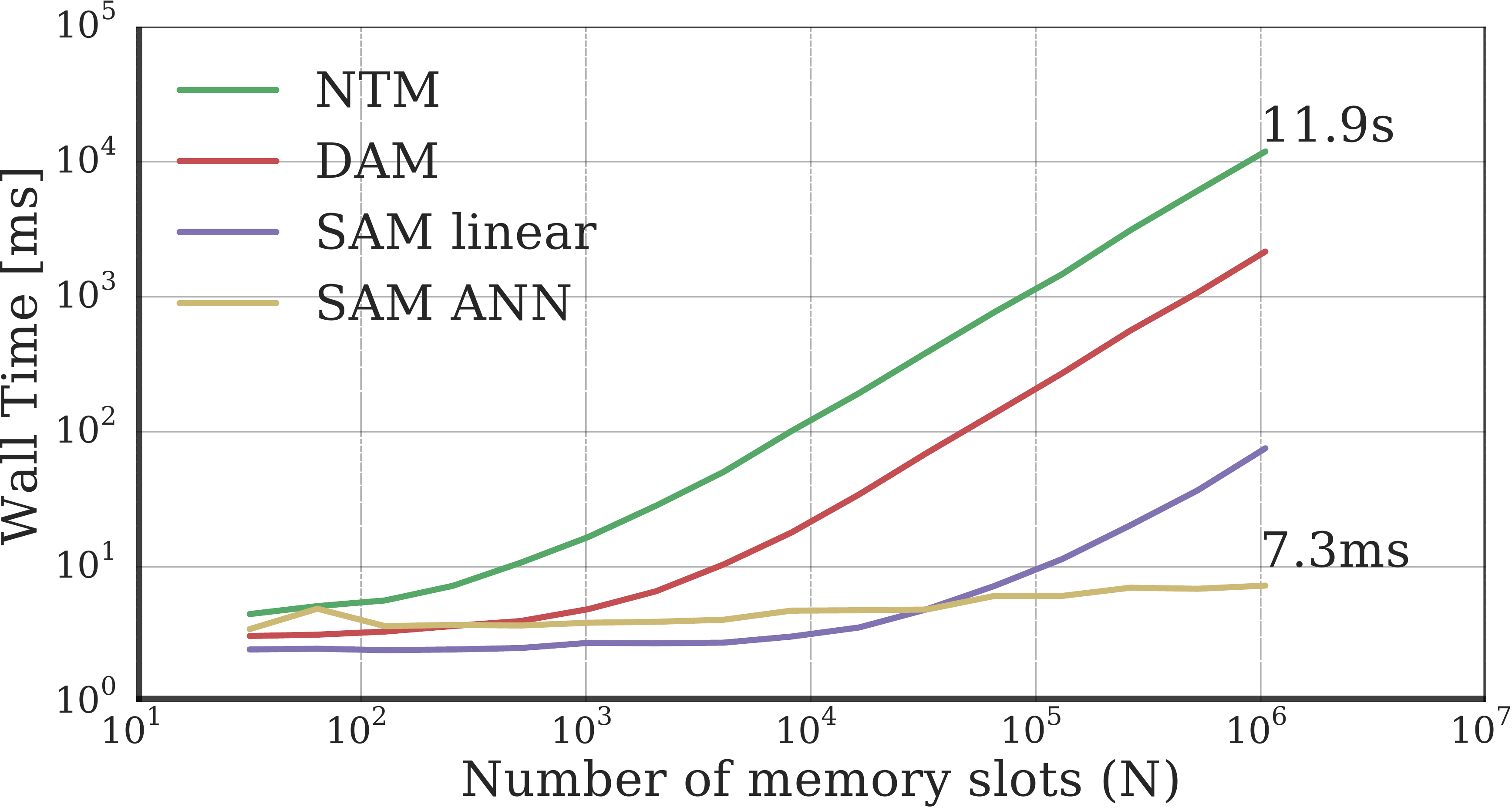}
    \caption{ \label{sf:speed}}
    \end{subfigure}
    %
    \begin{subfigure}{0.47\textwidth}
    \includegraphics[height=3.4cm]{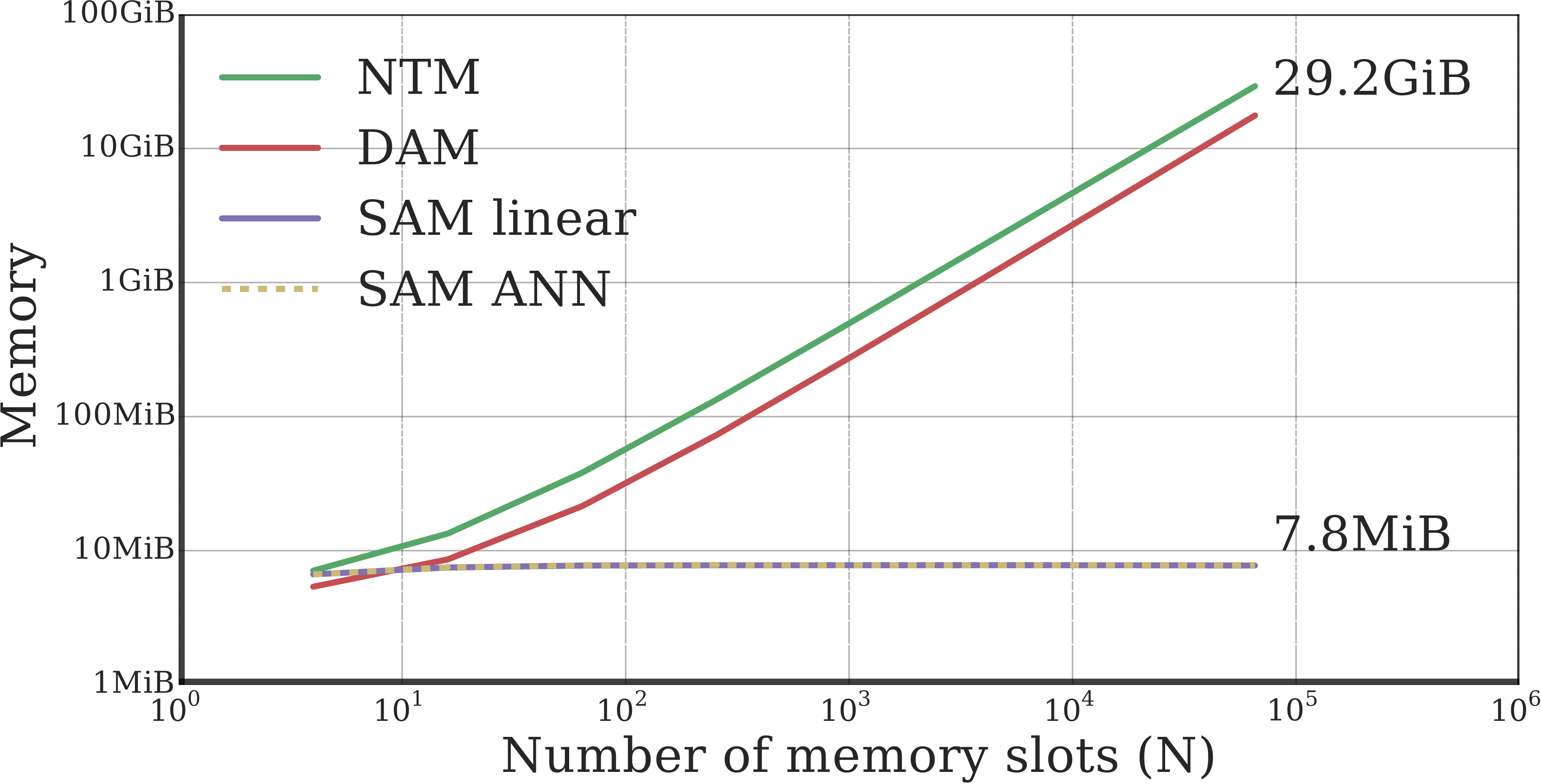}
    \caption{ \label{sf:memory}}
    \end{subfigure}
    \caption{\textbf{(\subref{sf:speed})} Wall-clock time of a single forward and backward pass. The k-d tree is a FLANN randomized ensemble with 4 trees and 32 checks. For 1M memories a single forward and backward pass takes $\SI{12}{s}$ for the NTM and $\SI{7}{ms}$ for SAM, a speedup of $1600\times$.
    \textbf{(\subref{sf:memory})} Memory used to train over sequence of 100 time steps, excluding initialization of external memory. The space overhead of SAM is independent of memory size, which we see by the flat line. When the memory contains 64,000 words the NTM consumes $\SI{29}{GiB}$ whereas SAM consumes only $\SI{7.8}{MiB}$, a compression ratio of $3700$.}
    \label{fig:perf_benchmarks}
\end{figure*}

We measured the forward and backward times of the SAM architecture versus the dense DAM variant and the original NTM (details of setup in Supplementary \ref{sec:benchmarking}). SAM is over $100$ times faster than the NTM when the memory contains one million words and an exact linear-index is used, and $1600$ times faster with the k-d tree (Figure \ref{fig:perf_benchmarks}\subref{sf:speed}). With an ANN the model runs in sublinear time with respect to the memory size. SAM's memory usage per time step is independent of the number of memory words (Figure \ref{fig:perf_benchmarks}\subref{sf:memory}), which empirically verifies the $\mathcal{O}(1)$ space claim from Supplementary \ref{sec:space_time_suppl}. For $\SI{64}{K}$ memory words SAM uses $\SI{53}{MiB}$ of physical memory to initialize the network and $\SI{7.8}{MiB}$ to run a 100 step forward and backward pass, compared with the NTM which consumes $\SI{29}{GiB}$.

\subsection{Learning with sparse memory access}

\label{ss:ntm_tasks}

We have established that SAM reaps a huge computational and memory advantage of previous models, but can we really learn with SAM's sparse approximations? We investigated the learning cost of inducing sparsity, and the effect of placing an approximate nearest neighbor index within the network, by comparing SAM with its dense variant DAM and some established models, the NTM and the LSTM.

We trained each model on three of the original NTM tasks \cite{graves2014neural}. \textbf{1. Copy}: copy a random input sequence of length 1--20, \textbf{2. Associative Recall}: given 3-6 random (key, value) pairs, and subsequently a cue key, return the associated value. \textbf{3. Priority Sort}: Given 20 random keys and priority values, return the top 16 keys in descending order of priority. We chose these tasks because the NTM is known to perform well on them.



\begin{figure*}
    \centering
    \begin{subfigure}{0.32\textwidth}
    \includegraphics[width=\linewidth]{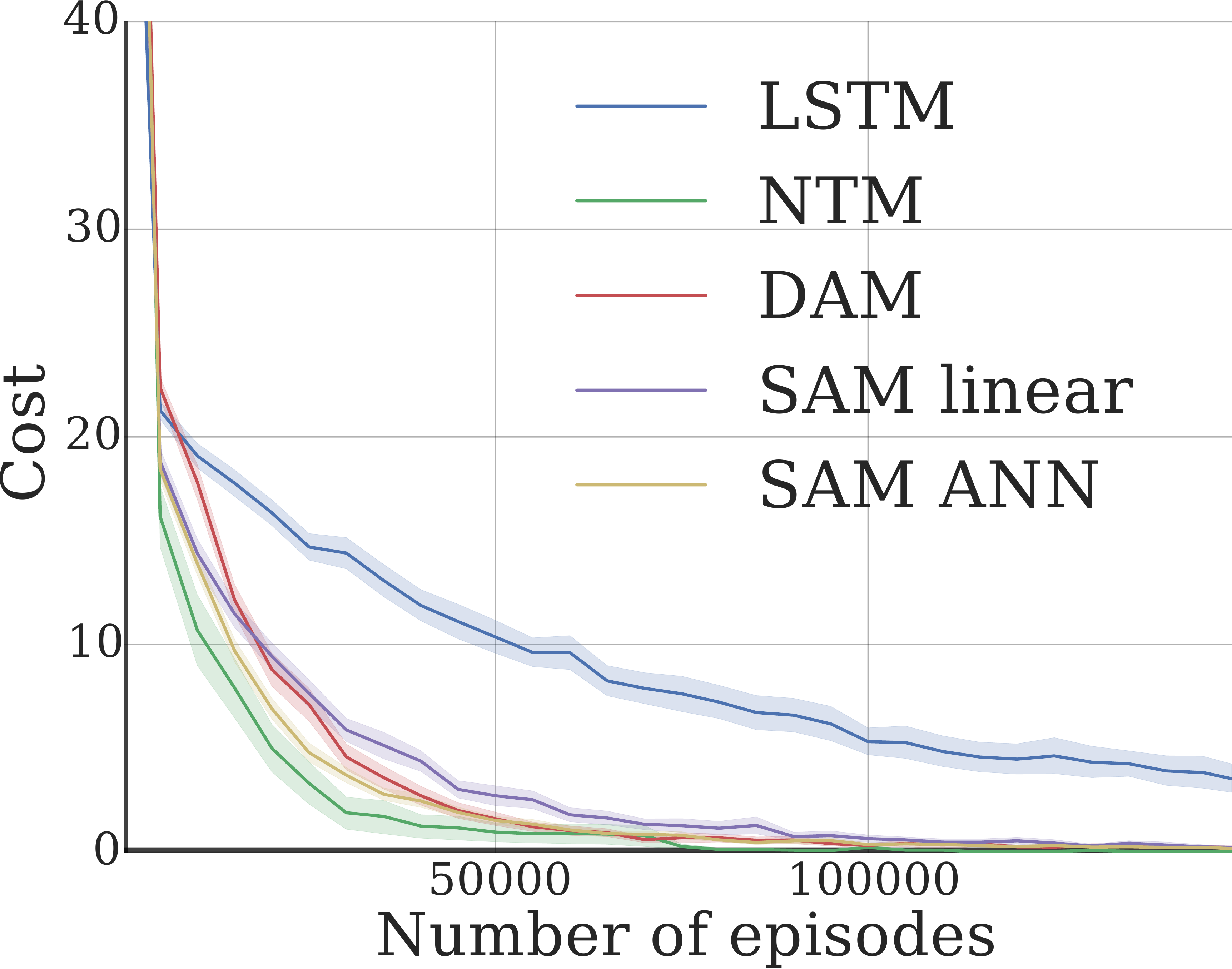}
    \caption{ \label{sf:ntm_copy} Copy}
    \end{subfigure}
    %
    \begin{subfigure}{0.32\textwidth}
    \includegraphics[width=\linewidth]{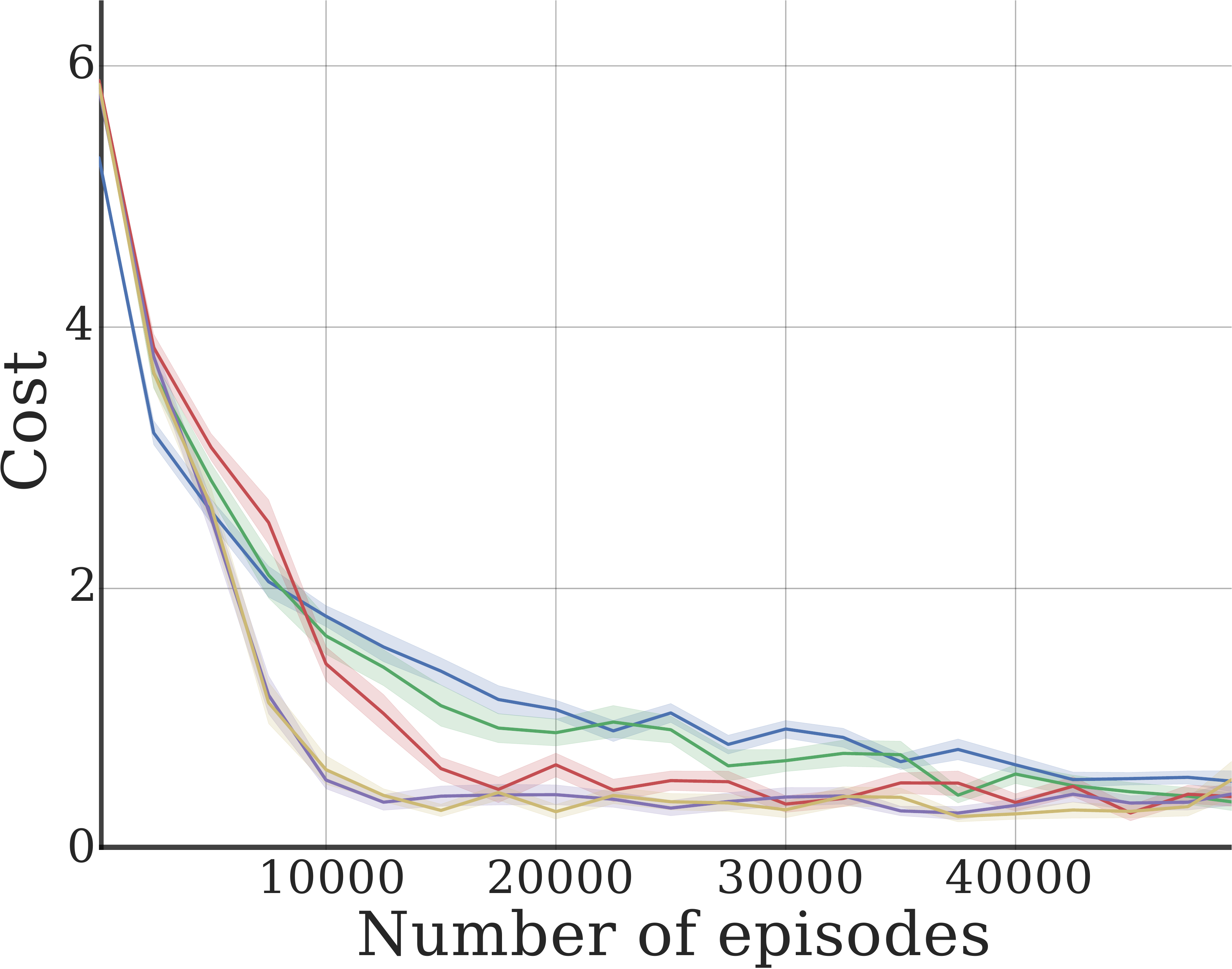}
    \caption{ \label{sf:ntm_associative_recall} Associative Recall}
    \end{subfigure}
    %
    %
    \begin{subfigure}{0.32\textwidth}
    \includegraphics[width=\linewidth]{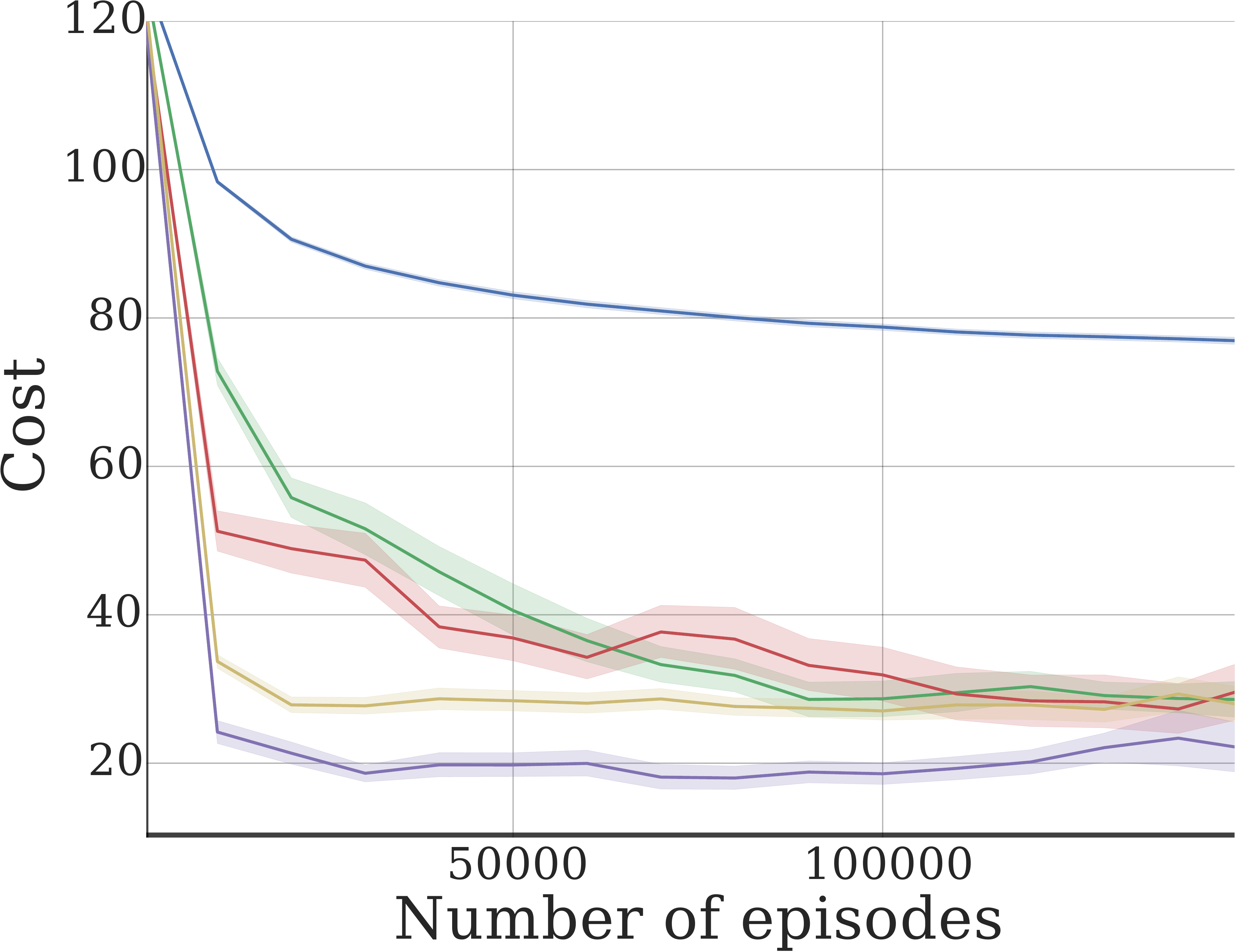}
    \caption{ \label{sf:ntm_topk} Priority Sort}
    \end{subfigure}
    \caption{Training curves for sparse (SAM) and dense (DAM, NTM) models. SAM trains comparably for the Copy task, and reaches asymptotic error significantly faster for Associative Recall and Priority Sort.
    Light colors indicate one standard deviation over 30 random seeds.}
    \label{fig:ntm_tasks}
\end{figure*}

Figure \ref{fig:ntm_tasks} shows that sparse models are able to learn with comparable efficiency to the dense models and, surprisingly, learn more effectively for some tasks --- notably priority sort and associative recall. This shows that sparse reads and writes can actually benefit early-stage learning in some cases.

Full hyperparameter details are in Supplementary \ref{sec:training_details}.


\subsection{Scaling with a curriculum}

The computational efficiency of SAM opens up the possibility of training on tasks that require storing a large amount of information over long sequences. Here we show this is possible in practice, by scaling tasks to a large scale via an exponentially increasing curriculum.

We parametrized three of the tasks described in Section \ref{ss:ntm_tasks}: associative recall, copy, and priority sort, with a progressively increasing difficulty level which characterises the length of the sequence and number of entries to store in memory. For example, level specifies the input sequence length for the copy task.
We exponentially increased the maximum level $h$ when the network begins to learn the fundamental algorithm.
Since the time taken for a forward and backward pass scales $\mathcal{O}(T)$ with the sequence length $T$, following a standard linearly increasing curriculum could potentially take $\mathcal{O}(T^2)$, if the same amount of training was required at each step of the curriculum.
Specifically, $h$ was doubled whenever the average training loss dropped below a threshold for a number of episodes. The level was sampled for each minibatch from the uniform distribution over integers $\mathcal{U}(0, h)$.

We compared the dense models, NTM and DAM, with both SAM with an exact nearest neighbor index (SAM linear) and with locality sensitive hashing (SAM ANN). The dense models contained 64 memory words, while the sparse models had $2 \times 10^6$ words. These sizes were chosen to ensure all models use approximately the same amount of physical memory when trained over 100 steps.


%

\begin{figure*}[h]
    \begin{minipage}[h]{0.32\textwidth}
        \centering
        \includegraphics[width=\linewidth]{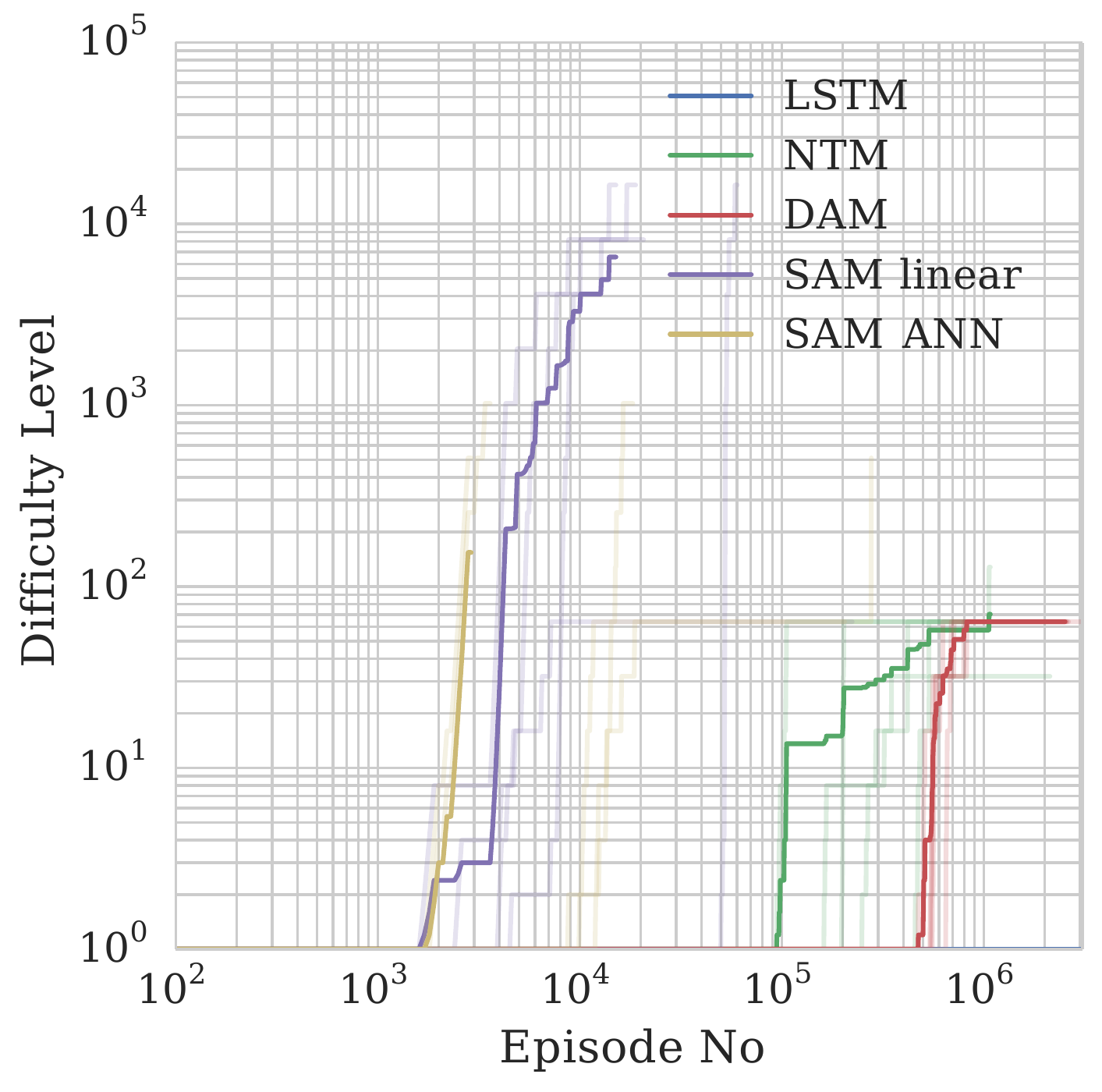}
        \subcaption{\label{sf:apc}}
    \end{minipage} %
    \begin{minipage}[h]{0.32\textwidth}
        \centering
        \includegraphics[width=\linewidth]{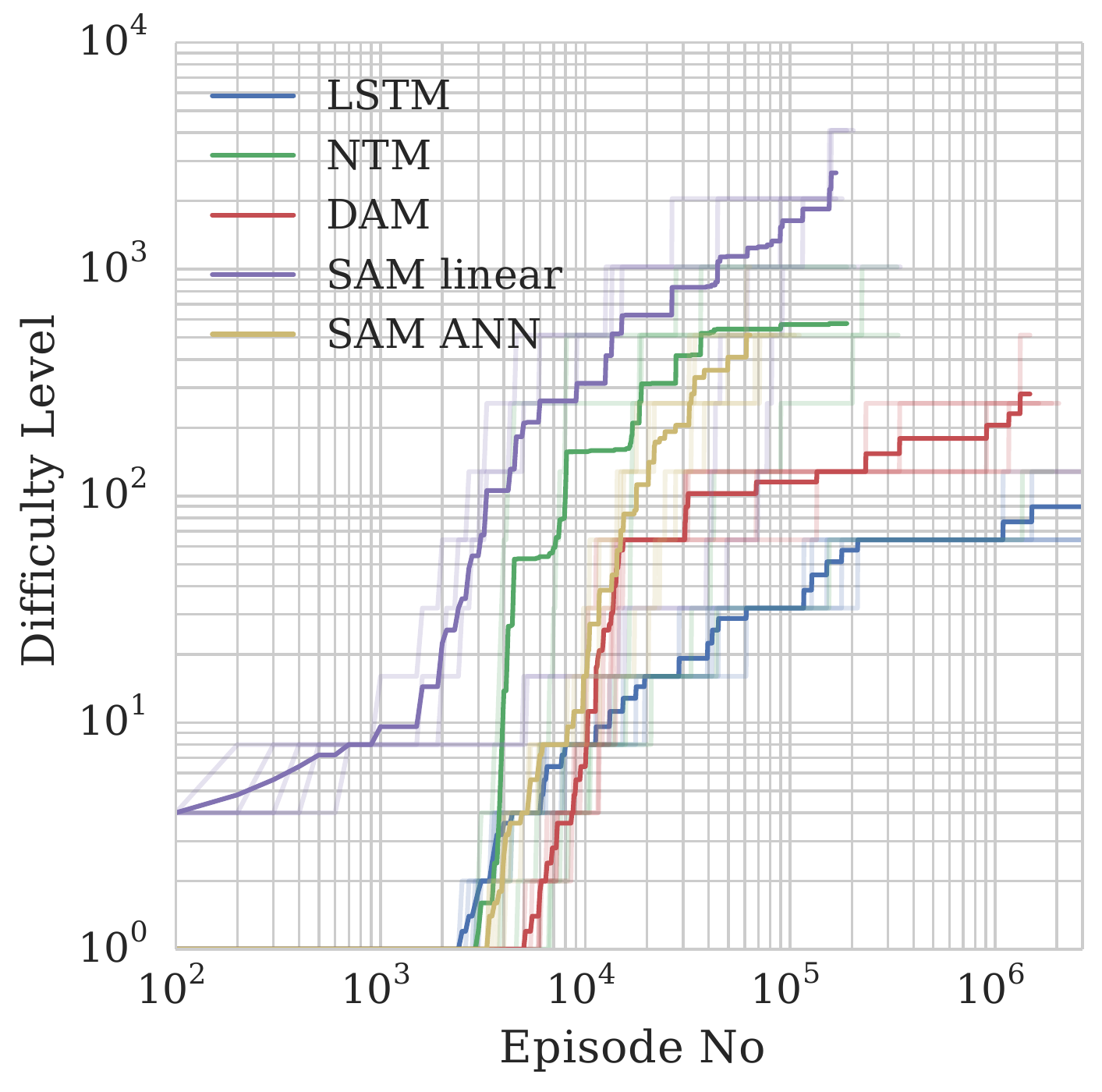}
        \subcaption{\label{sf:cc}}
    \end{minipage} %
    \begin{minipage}[h]{0.32\textwidth}
        \centering
        \includegraphics[width=\linewidth]{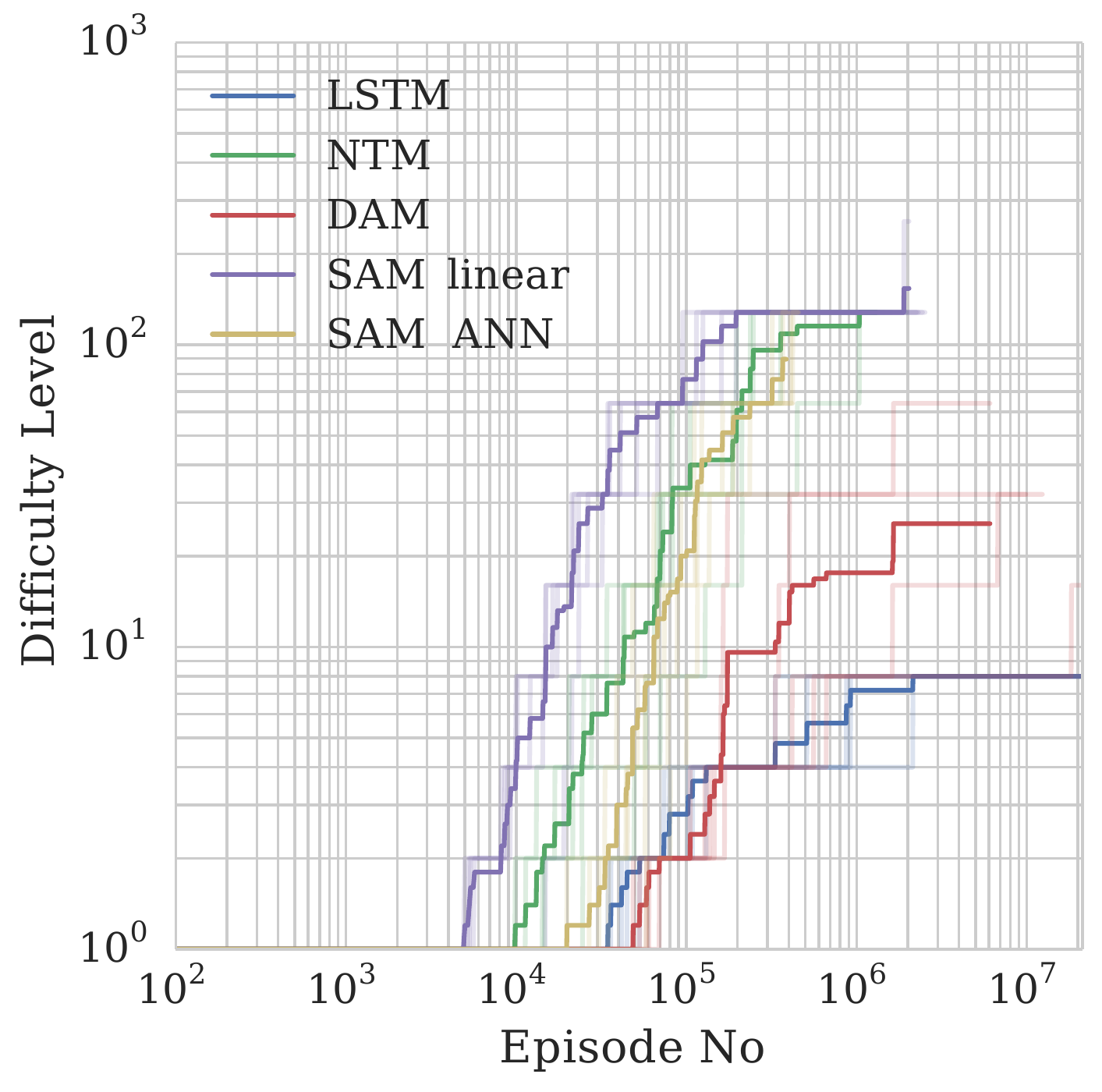}
        \subcaption{\label{sf:topkc}}
    \end{minipage}
    \\
    \caption{Curriculum training curves for sparse and dense models on (\subref{sf:apc}) Associative recall, (\subref{sf:cc}) Copy, and (\subref{sf:topkc}) Priority sort. Difficulty level indicates the task difficulty (e.g. the length of sequence for copy). We see SAM train (and backpropagate over) episodes with thousands of steps, and tasks which require thousands of words to be stored to memory. Each model is averaged across 5 replicas of identical hyper-parameters (light lines indicate individual runs).
    }
%
    \label{fig:curriculumT}
\end{figure*}

For all tasks, SAM was able to advance further than the other models, and in the associative recall task, SAM was able to advance through the curriculum to sequences greater than $4000$ (Figure \ref{fig:curriculumT}). Note that we did not use truncated backpropagation, so this involved BPTT for over $4000$ steps with a memory size in the millions of words.


To investigate whether SAM was able to learn algorithmic solutions to tasks, we investigated its ability to generalize to sequences that far exceeded those observed during training. Namely we trained SAM on the associative recall task up to sequences of length $10,000$, and found it was then able to generalize to sequences of length $200,\!000$ (Supplementary Figure  \ref{fig:extrapolation}). 





\subsection{Question answering on the Babi tasks}

\cite{weston2015towards} introduced toy tasks they considered a prerequisite to agents which can reason and understand natural language. They are synthetically generated language tasks with a vocab of about 150 words that test various aspects of simple reasoning such as deduction, induction and coreferencing.

We tested the models (including the Sparse Differentiable Neural Computer described in Supplementary \ref{sec:sdnc}) on this task. The full results and training details are described in Supplementary \ref{sec:babi}.

The MANNs, except the NTM, are able to learn solutions comparable to the previous best results, failing at only 2 of the tasks. The SDNC manages to solve all but 1 of the tasks, the best reported result on Babi that we are aware of.

Notably the best prior results have been obtained by using supervising the memory retrieval (during training the model is provided annotations which indicate which memories should be used to answer a query). More directly comparable previous work with end-to-end memory networks, which did not use supervision \cite{sukhbaatar2015end}, fails at 6 of the tasks.

Both the sparse and dense perform comparably at this task, again indicating the sparse approximations do not impair learning. We believe the NTM may perform poorly since it lacks a mechanism which allows it to allocate memory effectively.


\subsection{Learning on real world data}

Finally, we demonstrate that the model is capable of learning in a non-synthetic dataset. Omniglot \cite{lake2015human} is a dataset of 1623 characters taken from 50 different alphabets,
with 20 examples of each character.
This dataset is used to test rapid, or \textit{one-shot} learning, since there are few examples of each character but many different character classes. Following \cite{santoro2016}, we generate episodes where a subset of characters are randomly selected from the dataset, rotated and stretched, and assigned a randomly chosen label. At each time step an example of one of the characters is presented, along with the correct label of the proceeding character. Each character is presented 10 times in an episode (but each presentation may be any one of the 20 examples of the character). In order to succeed at the task the model must learn to rapidly associate a novel character with the correct label, such that it can correctly classify subsequent examples of the same character class.

\begin{figure}[h]
  \includegraphics[width=6cm]{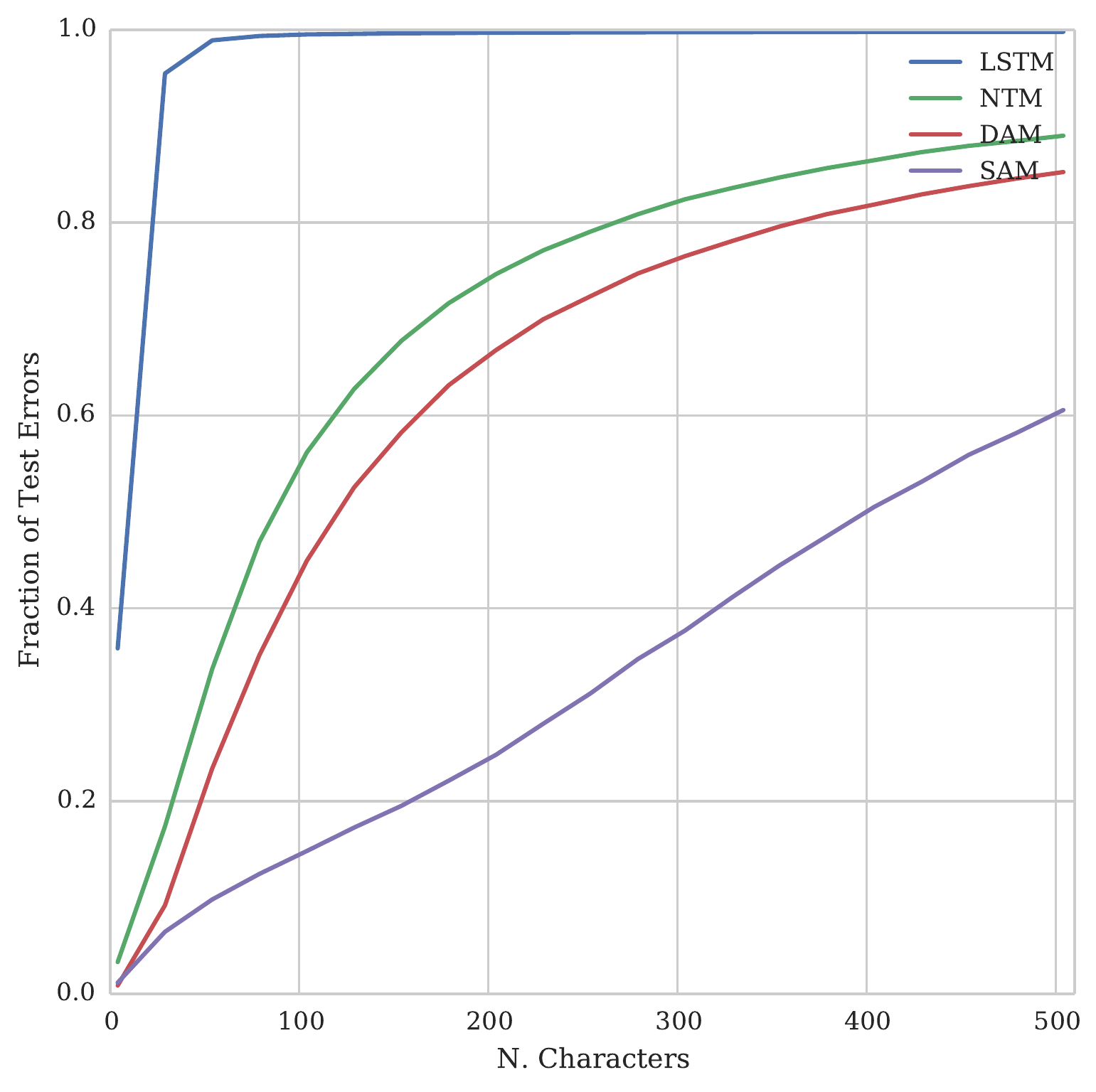}
  \caption{Test errors for the Omniglot task (described in the text) for the best runs (as chosen by the validation set). The characters used in the test set were not used in validation or training. All of the MANNs were able to perform much better than chance with $\approx 500$ characters (sequence lengths of $\approx 5000$), even though they were trained, at most, on sequences of $\approx 130$ (chance is $0.002$ for $500$ characters). This indicates they are learning generalizable solutions to the task. SAM is able to outperform other approaches, presumably because it can utilize a much larger memory. }
  \label{fig:omniglot}
\end{figure}

Again, we used an exponential curriculum, doubling the number of additional characters provided to the model whenever the cost was reduced under a threshold. After training all MANNs for the same length of time, a validation task with $500$ characters was used to select the best run, and this was then tested on a test set, containing all novel characters for different sequence lengths (Figure \ref{fig:omniglot}). All of the MANNs were able to perform much better than chance, even on sequences $\approx 4\times$ longer than seen during training. SAM outperformed other models, presumably due to its much larger memory capacity. Previous results on the Omniglot curriculum \cite{santoro2016} task are not identical, since we used 1-hot labels throughout and the training curriculum scaled to longer sequences, but our results with the dense models are comparable ($\approx 0.4$ errors with $100$ characters), while the SAM is significantly better ($0.2 <$ errors with $100$ characters).


%
%
%
%
%
%
%





\section{Discussion}

Scaling memory systems is a pressing research direction due to potential for compelling applications with large amounts of memory. We have demonstrated that you can train neural networks with large memories via a sparse read and write scheme that makes use of efficient data structures within the network, and obtain significant speedups during training. Although we have focused on a specific MANN (SAM), which is closely related to the NTM, the approach taken here is general and can be applied to many differentiable memory architectures, such as Memory Networks \cite{weston2014memory}.


It should be noted that there are multiple possible routes toward scalable memory architectures. For example, prior work aimed at scaling Neural Turing Machines \cite{zaremba2015reinforcement} used reinforcement learning to train a discrete addressing policy. This approach also touches only a sparse set of memories at each time step, but relies on higher variance estimates of the gradient during optimization. Though we can only guess at what class of memory models will become staple in machine learning systems of the future, we argue in Supplementary \ref{sec:space_time_suppl} that they will be no more efficient than SAM in space and time complexity if they address memories based on content.




We have experimented with randomized k-d trees and LSH within the network to reduce the forward pass of training to sublinear time, but there may be room for improvement here. K-d trees were not designed specifically for fully online scenarios, and can become imbalanced during training. Recent work in tree ensemble models, such as Mondrian forests \cite{lakshminarayanan2014mondrian}, show promising results in maintaining balanced hierarchical set coverage in the online setting. An alternative approach which may be well-suited is LSH forests \cite{bawa2005lsh}, which adaptively modifies the number of hashes used. It would be an interesting empirical investigation to more fully assess different ANN approaches in the challenging context of training a neural network.

Humans are able to retain a large, task-dependent set of memories obtained in one pass with a surprising amount of fidelity \cite{brady2008visual}. Here we have demonstrated architectures that may one day compete with humans at these kinds of tasks.






\subsection*{Acknowledgements}
We thank Vyacheslav Egorov, Edward Grefenstette, Malcolm Reynolds, Fumin Wang and Yori Zwols for their assistance, and the Google DeepMind family for helpful discussions and encouragement.

\small
\bibliography{example_paper}{}
\bibliographystyle{plain}

\newpage
\appendix

\section*{Supplementary Information}

\section{Time and space complexity}
\label{sec:space_time_suppl}

Under a reasonable class of content addressable memory architectures $\mathcal{A}$, SAM is optimal in time and space complexity.

\begin{definition}
\label{label:def_cam}
Let $\mathcal{M}$ be a collection of real vectors $m_1, m_2, \ldots, m_N$ of fixed dimension $d$. Let $\mathcal{A}$ be the set of all content addressable memory data structures that store $\mathcal{M}$ and can return at least one word $m_j$ such that $D(q, m_j) \le c \, (1 + \epsilon) $ for a given $L^p$ norm $D$, query vector $q$, and $\epsilon > 0$; provided such a memory $m_c \,$ exists with $D(q, m_c) = c$.
\end{definition}

Existing lower bounds \cite{motwani2007lower,arya1998optimal} assert that for any data structure $a \in \mathcal{A}$, $a$  requires $\Omega(\log N)$ time and $\Omega(N)$ space to perform a read operation. The SAM memory architecture proposed in this paper is contained within $\mathcal{A}$ as it computes the approximate nearest neighbors problem in fixed dimensions \cite{flann_pami_2014}. As we will show, SAM requires $\mathcal{O}(\log N)$ time to query and maintain the ANN, $\mathcal{O}(1)$ to perform all subsequent sparse read, write, and error gradient calculations. It requires $\mathcal{O}(N)$ space to initialize the memory and $\mathcal{O}(1)$ to store intermediate sparse tensors. We thus conclude it is optimal in asymptotic time and space complexity.

\subsection{Initialization}
Upon initialization, SAM consumes $\mathcal{O}(N)$ space and time to instantiate the memory and the memory Jacobian. Furthermore, it requires $\mathcal{O}(N)$ time and space to initialize auxiliary data structures which index the memory, such as the approximate nearest neighbor which provides a content-structured view of the memory, and the least accessed ring, which maintains the temporal ordering in which memory words are accessed. These initializations represent an unavoidable one-off cost that does not recur per step of training, and ultimately has little effect on training speed. For the remainder of the analysis we will concentrate on the space and time cost per training step.
\subsection{Read}
Recall the sparse read operation,

\begin{equation}
\label{eq:sparse_read_suppl}
\tilde r_t = \sum_{i = 1}^K   \tilde w_t^R(s_i) \mathbf{M}_t (s_i) \; .
\end{equation}

As $K$ is chosen to be a fixed constant, it is clear we can compute (\ref{eq:sparse_read_suppl}) in $\mathcal{O}(1)$ time. During the backward pass, we see the gradients are sparse with only $K$ non-zero terms,
\[
\frac{\partial L}{\partial \tilde  w^R_t}(i) =
\left \{
    \begin{array}{ll}
        \mathbf{M}_t(i) \cdot \frac{\partial L}{\partial \tilde r_t} & \hbox{if } i \in \{s_1, s_2, \ldots, s_K\} \\
        0 & \hbox{otherwise.}
    \end{array}
\right .
\]
and
\[
\frac{\partial L}{\partial  M_t}(i) =
\left \{
    \begin{array}{ll}
        \tilde w^R_t(i) \frac{\partial L}{\partial \tilde r_t} & \hbox{if } i \in \{s_1, s_2, \ldots, s_K\} \\
        \mathbf{0} & \hbox{otherwise.}
    \end{array}
\right .
\]
where $\mathbf{0}$ is a vector of $M$ zeros. Thus they can both be computed in constant time by skipping the computation of zeros. Furthermore by using an efficient sparse matrix format to store these matrices and vectors, such as the CSR, we can represent them using at most $3K$ values. Since the read word $\tilde r_t$ and its respective error gradient is the size of a single word in memory ($M$ elements), the overall space complexity is $\mathcal{O}(1)$ per time step for the read.

\subsection{Write}
Recall the write operation,

\begin{equation}
    \label{eq:sparse_write_suppl}
    \mathbf{M}_t \leftarrow \mathbf{M}_{t - 1} - \mathbf{E}_t  + \mathbf{A}_t, \; ,
\end{equation}

where $\mathbf{A}_t = w^{\W}_t a_t^T$ is the add matrix, $\mathbf{E_t} = \mathbf{M}_{t - 1} \odot \mathbf{R}_t$ is the erase matrix, and $\mathbf{R}_t = \mathbb{I}^U_t \mathbf{1}^T$ is defined to be the erase weight matrix. We chose the write weights to be an interpolation between the least recently accessed location and the previously read locations,

\begin{equation}
    \label{eq:lru_write_weights_suppl}
    w^{\W}_t = \alpha_t \, \left( \gamma_t \, \tilde w^R_{t-1} + (1 - \gamma_t) \, \mathbb{I}^U_t \right) \, .
\end{equation}

For sparse reads where $w^{R}_t = \tilde w^{R}_t$ is a sparse vector with $K$ non-zeros, the write weights $w^{\W}_t$ is also sparse with $K + 1$ non-zeros: $1$ for the least recently accessed location and $K$ for the previously read locations. Thus the sparse-dense outer product $\mathbf{A_t} = w^{\W}_t a_t^T$ can be performed in $\mathcal{O}(1)$ time as $K$ is a fixed constant.

Since $\mathbf{R}_t = \mathbb{I}^U_t \mathbf{1}^T$ can be represented as a sparse matrix with one single non-zero, the erase matrix $\mathbf{E}_t$ can also. As $\mathbf{A_t}$ and $\mathbf{E_t}$ are sparse matrices we can then add them component-wise to the dense $\mathbf{M_{t-1}}$ in $\mathcal{O}(1)$ time. By analogous arguments the backward pass can be computed in $\mathcal{O}(1)$ time and each sparse matrix can be represented in $\mathcal{O}(1)$ space.

\begin{figure}[H]
    \centering
    \includegraphics[width=\columnwidth]{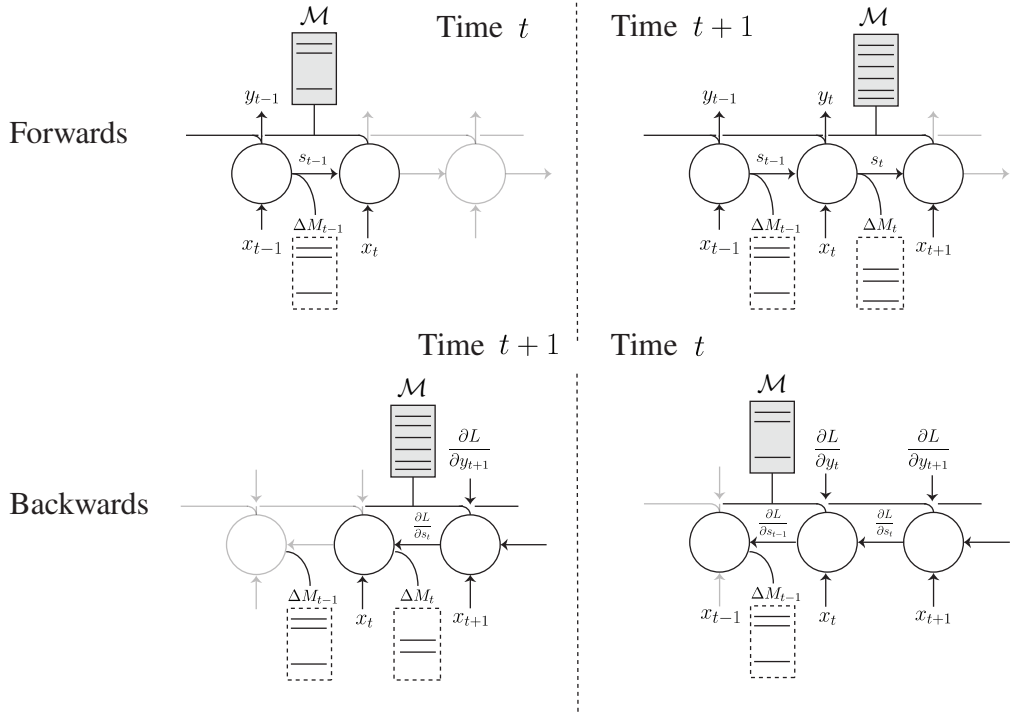}
    \caption{A schematic of the memory efficient backpropagation through time. Each circle represents an instance of the SAM core at a given time step. The grey box marks the dense memory. Each core holds a reference to the single instance of the memory, and this is represented by the solid connecting line above each core. We see during the forward pass, the memory's contents are modified sparsely, represented by the solid horizontal lines. Instead of caching the changing memory state, we store only the sparse modifications --- represented by the dashed white boxes. During the backward pass, we ``revert'' the cached modifications to restore the memory to its prior state, which is crucial for correct gradient calculations.}
    \label{fig:bptt}
\end{figure}

We avoid caching the modified memory, and thus duplicating it, by applying the write directly to the memory. To restore its prior state during the backward pass, which is crucial to gradient calculations at earlier time steps, we roll the memory it back by reverting the sparse modifications with an additional $\mathcal{O}(1)$ time overhead (Supplementary Figure \ref{fig:bptt}).

The location of the least recently accessed memory can be maintained in $\mathcal{O}(1)$ time by constructing a circular linked list that tracks the indices of words in memory, and preserves a strict ordering of relative temporal access. The first element in the ring is the least recently accessed word in memory, and the last element in the ring is the most recently modified. We keep a ``head'' pointer to the first element in the ring. When a memory word is randomly accessed, we can push its respective index to the back of the ring in $\mathcal{O}(1)$ time by redirecting a small number of pointers. When we wish to pop the least recently accessed memory (and write to it) we move the head to the next element in the ring in $\mathcal{O}(1)$ time.

\subsection{Content-based addressing}
As discussed in Section \ref{sec:ann} we can calculate the content-based attention, or read weights $w^{R}_t$, in $\mathcal{O}(\log N)$ time using an approximate nearest neighbor index that views the memory. We keep the ANN index synchronized with the memory by passing it through the network as a non-differentiable member of the network's state (so we do not pass gradients for it), and we update the index upon each write or erase to memory in $\mathcal{O}(\log N)$ time. Maintaining and querying the ANN index represents the most expensive part of the network, which is reasonable as content-based addressing is inherently expensive \cite{motwani2007lower,arya1998optimal}.

For the backward pass computation, specifically calculating $\frac{\partial L}{\partial q_t}$ and $\frac{\partial L}{\partial \mathbf{M_t}}$ with respect to $w^R_t$, we can once again compute these using sparse matrix operations in $\mathcal{O}(1)$ time. This is because the $K$ non-zero locations have been determined during the forward pass.

Thus to conclude, SAM consumes in total $\mathcal{O}(1)$ space for both the forward and backward step during training, $\mathcal{O}(\log N)$ time per forward step, and $\mathcal{O}(1)$ per backward step.

\section{Control flow}

\begin{figure}[H]
    \centering
    \includegraphics[scale=0.7]{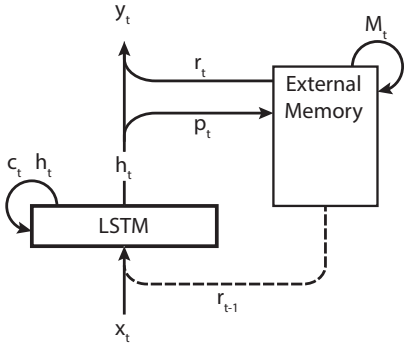}
    \caption{Schematic showing how the controller interfaces with the external memory in our experiments. The controller (LSTM) output $h_t$ is used (through a linear projection, $p_t$) to read and write to the memory. The result of the read operation $r_t$ is combined with $h_t$ to produce output $y_t$, as well as being feed into the controller at the next timestep ($r_{t-1}$).}
    \label{fig:controlflow}
\end{figure}

\section{Training details}
\label{sec:training_details}
Here we provide additional details on the training regime used for our experiments used in Figure \ref{fig:ntm_tasks}.

To avoid bias in our results, we chose the learning rate that worked best for DAM (and not SAM). We tried learning rates $\{10^{-6}, 5 \times 10^{-5}, 10^{-5}, 5 \times 10^{-4}, 10^{-4}\}$ and found that DAM trained best with $10^{-5}$. We also tried values of $K$ $\{4, 8, 16\}$ and found no significant difference in performance across the values.  We used 100 hidden units for the LSTM (including the controller LSTMs), a minibatch of $8$, $8$ asynchronous workers to speed up training, and \texttt{RMSProp} \cite{tieleman2012lecture} to optimize the controller. We used $4$ memory access heads and configured SAM to read from only $K = 4$ locations per head.

\section{Sparse Differentiable Neural Computer}
\label{sec:sdnc}

Recently \cite{graves2016dnc} proposed a novel MANN the Differentiable Neural Computer (DNC). The two innovations proposed by this model are a new approach to tracking memory freeness (dynamic memory allocation) and a mechanism for associating memories together (temporal memory linkage).
We demonstrate here that the approaches enumerated in the paper can be adapted to new models by outlining a sparse version of this model, the Sparse Differentiable Neural Computer (SDNC), which learns with similar data efficiency while retaining the computational advantages of sparsity.

\subsection{Architecture}
For brevity, we will only explain the sparse implementations of these two items, for the full model details refer to the original paper. The mechanism for sparse memory reads and writes was implemented identically to SAM.

It is possible to implement a scalable version of the dynamic memory allocation system of the DNC avoiding any $O(N)$ operations by using a heap. However, because it is practical to run the SDNC with many more memory words, reusing memory is less crucial so we did not implement this and used the same usage tracking as in SAM.

The temporal memory linkage in the DNC is a system for associating and recalling memory locations which were written in a temporal order, for exampling storing and retrieving a list. In the DNC this is done by maintaining a temporal linkage matrix $\mathbf{L}_t \in [0, 1]^{N \times N}$. $\mathbf{L}_t[i,j]$ represents the degree to which location $i$ was written to after location $j$. This matrix is updated by tracking the precedence weighting $p_t$, where $p_t(i)$ represents the degree to which location $i$ was written to.
\begin{align}
    p_0 &= 0    \\
    p_t &= (1 - \sum_i w_t^{W}(i) ) \, p_{t-1} + w_t^{W}
\end{align}
The memory linkage is updated according to the following recurrence
\begin{align}
\mathbf{L}_0 &= 0 \\
\mathbf{L}_t(i, j) &= \left \{
                \begin{array}{ll}
                    0 & i = j  \\
                    (1 - w_t^{W}(i) - w_t^{W}(j)) \mathbf{L}_{t -1}(i, j) + w_t^{W}(i) p_{t-1}(j) & i \neq j \\
                \end{array} \right . \\
\end{align}

The temporal linkage $\mathbf{L}_t$ can be used to compute read weights following the temporal links either forward
\begin{equation}
f_t^r = \mathbf{L}_t w^{r}_{t-1}
\end{equation}
or backward
\begin{equation}
b_t^r = \mathbf{L}_t^{T} w^{r}_{t-1}
\end{equation}
The read head then uses a 3-way softmax to select between a content-based read or following the forward or backward weighting.

Naively, the link matrix requires $O(N^2)$ memory and computation although \cite{graves2016dnc} proposes a method to reduce the computational cost to $O(N \log N)$ and $O(N)$ memory cost.

In order to maintain the scaling properties of the SAM, we wish to avoid any computational dependence on $N$. We do this by maintaining two sparse matrices $\mathbf{N}_t, \mathbf{P}_t \in [0, 1]^{N \times \KL}$ that approximate $\mathbf{L}_t$ and $\mathbf{L}^T_t$ respectively.  We store these matrices in Compressed Sparse Row format. They are defined by the following updates:
\begin{align}
    \mathbf{N}_0 &= 0 \\
    \mathbf{P}_0 &= 0 \\
    \mathbf{N}_t(i, j) &= (1 - w_t^W(i) ) \, \mathbf{N}_{t-1}(i, j) + w_t^W(i)\; p_{t-1}(j) \\
    \mathbf{P}_t(i, j) &= (1 - w_t^W(j) ) \, \mathbf{P}_{t-1}(i, j) + w_t^W(j)\; p_{t-1}(i)
\end{align}
Additionally, $p_t$ is, as with the other weight vectors maintained as a sparse vector with at most $K_L$ non-zero entries. This means that the outer product of $w_t p_{t-1}^T$ has at most $K_L^2$ non-zero entries. In addition to the updates specified above, we also constrain each row of the matrices $\mathbf{N}_t$ and $\mathbf{P}_t$ to have at most $K_L$ non-zero entries --- this constraint can be applied in $O(K_L^2)$ because at most $K_L$ rows change in the matrix.

Once these matrices are applied the read weights following the temporal links can be computed similar to before:
\begin{align}
f_t^r &= \mathbf{N}_t w^{r}_{t-1} \\
b_t^r &= \mathbf{P}_t w^{r}_{t-1}
\end{align}

Note, the number of locations we read from, $K$, does not have to equal the number of outward and inward links we preserve, $K_L$. We typically choose $K_L = 8$ as this is still very fast to compute ($100 \mu s$ in total to calculate $\mathbf{N}_t, \mathbf{P}_t, p_t, f_t^r, b_t^r$ on a single CPU thread) and we see no learning benefit with larger $K_L$. In order to compute the gradients, $\mathbf{N}_t$ and $\mathbf{P}_t$ need to be stored. This could be done by maintaining a sparse record of the updates applied and reversing them, similar to that performed with the memory as described in Section \ref{sec:efficient_bptt}. However, for implementation simplicity we did not pass gradients through the temporal linkage matrices.

\subsection{Results}
\begin{figure*}[h]
    \centering
    \begin{subfigure}{0.47\textwidth}
    \includegraphics[height=3.4cm]{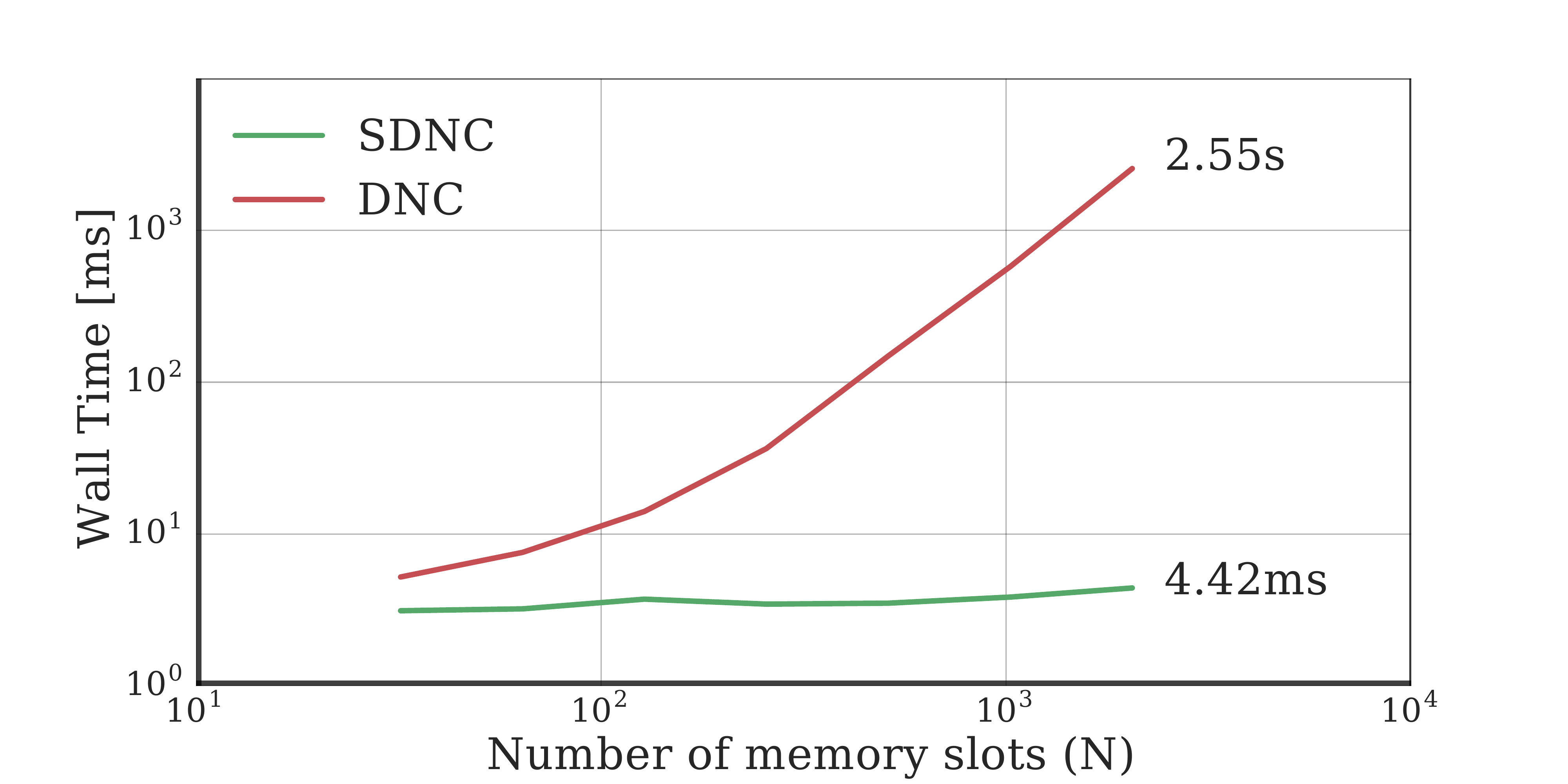}
    \caption{ \label{sf:dnc_speed}}
    \end{subfigure}
    %
    \begin{subfigure}{0.47\textwidth}
    \includegraphics[height=3.4cm]{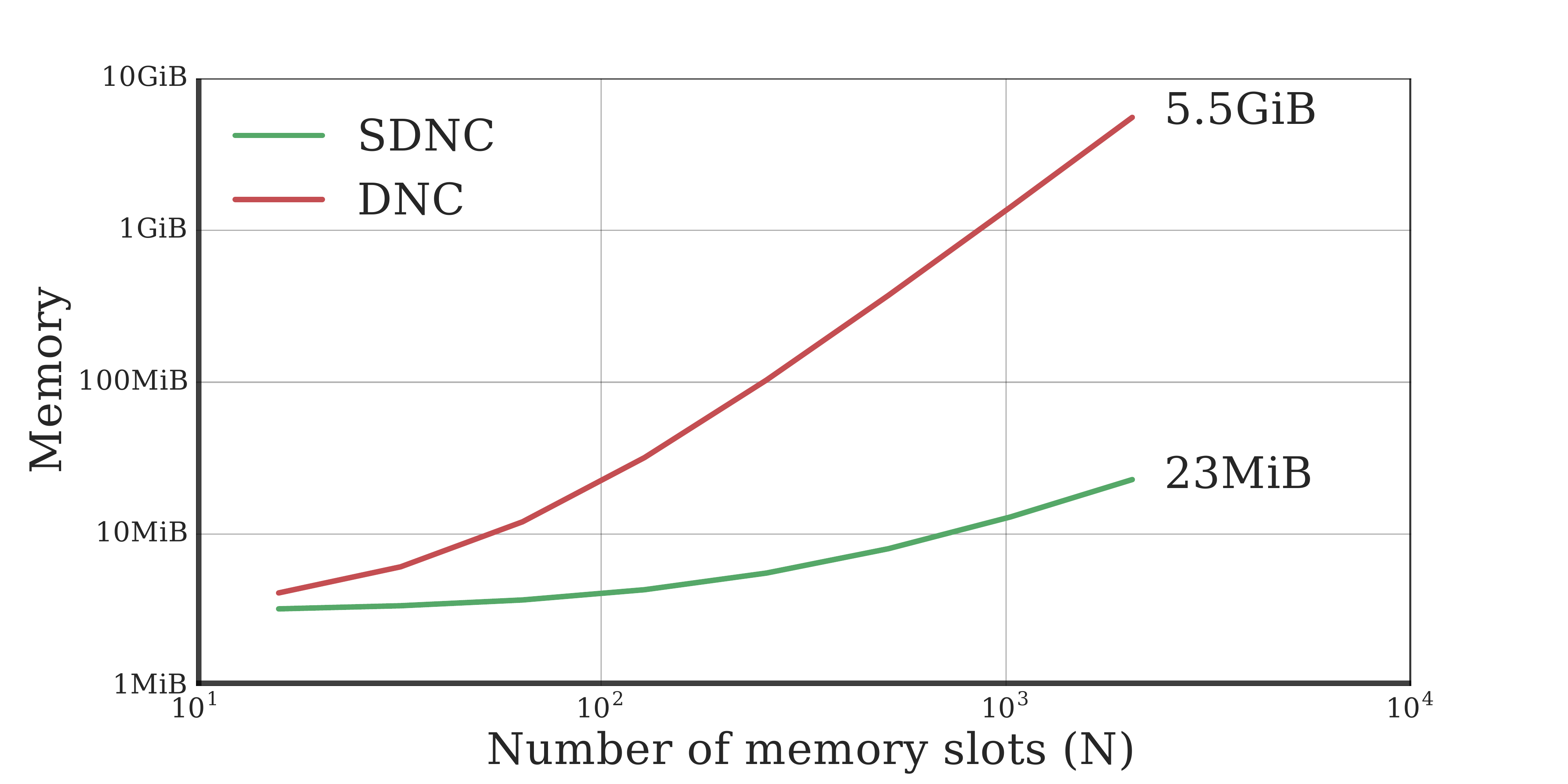}
    \caption{ \label{sf:dnc_memory}}
    \end{subfigure}
    \caption{Performance benchmarks between the DNC and SDNC for small to medium memory sizes. Here the SDNC uses a linear KNN. \textbf{(\subref{sf:dnc_speed})} Wall-clock time of a single forward and backward pass.
    \textbf{(\subref{sf:dnc_memory})} Total memory usage (including initialization) when trained over sequence of $10$ time steps.
    }
    \label{fig:dnc_perf_benchmarks}
\end{figure*}

We benchmarked the speed and memory performance of the SDNC versus a naive DNC implementation (details of setup in Supplementary \ref{sec:benchmarking}). The results are displayed in Figure \ref{fig:dnc_perf_benchmarks}. Here, the computational benefits of sparsity are more pronounced due to the expensive (quadratic time and space) temporal transition table operations in the DNC. We were only able to run comparative benchmarks up to $N = 2048$, as the DNC quickly exceeded the machine's physical memory for larger values; however even at this modest memory size we see a speed increase of $\approx 440 \times$ and physical memory reduction of $\approx 240 \times$. Note, unlike the SAM memory benchmark in Section \ref{sec:results} we plot the total memory consumption, i.e. the memory overhead of the initial start state plus the memory overhead of unrolling the core over a sequence. This is because the SDNC and DNC do not have identical start states. The sparse temporal transition matrices $\mathbf{N}_0, \mathbf{P}_0 \in [0, 1]^{N \times N\K}$ consume much less memory than the corresponding $\mathbf{L}_0 \in [0, 1]^{N \times N}$ in the DNC.

In order to compare the models on an interesting task we ran the DNC and SDNC on the Babi task (this task is described more fully in the main text). The results are described in Supplementary \ref{sec:babi} and demonstrate the SDNC is capable of learning competitively. In particular, it achieves the best report result on the Babi task.

\section{Benchmarking details}
\label{sec:benchmarking}

Each model contained an LSTM controller with 100 hidden units, an external memory containing $N$ slots of memory, with word size $32$ and $4$ access heads. For speed benchmarks, a minibatch size of $8$ was used to ensure fair comparison - as many dense operations (e.g. matrix multiplication) can be batched efficiently. For memory benchmarks, the minibatch size was set to $1$.

We used Torch7 \cite{collobert2011torch7} to implement SAM, DAM, NTM, DNC and SDNC. Eigen v3 \cite{guennebaud2010eigen} was used for the fast sparse tensor operations, using the provided CSC and CSR formats. All benchmarks were run on a Linux desktop running Ubuntu 14.04.1 with 32GiB of RAM and an Intel Xeon E5-1650 3.20GHz processor with power scaling disabled.

\section{Generalization on associative recall}

\begin{figure}[h!]
    \centering
    \includegraphics[width=0.6\linewidth]{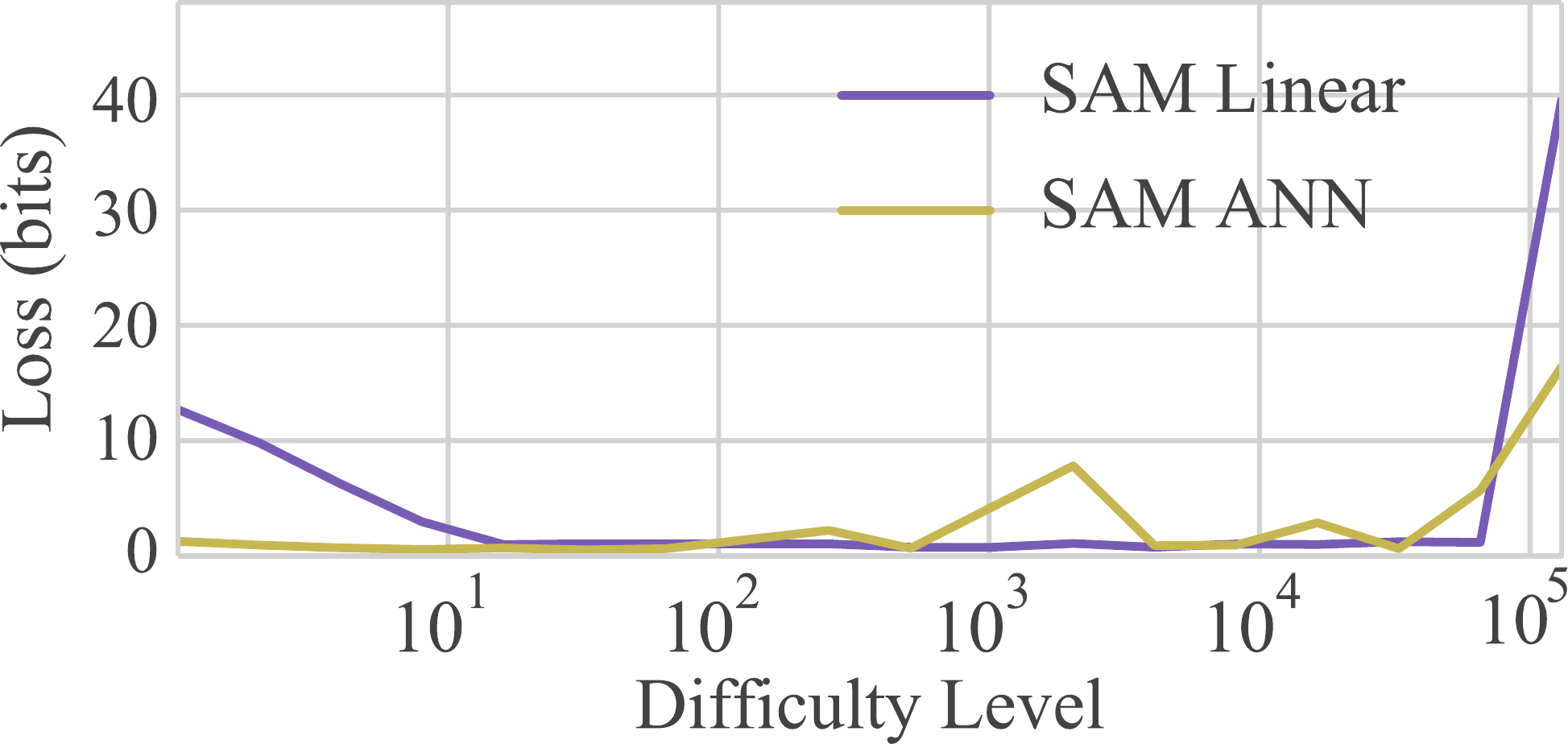}
    \caption{We tested the generalization of SAM on the associative recall task. We train each model up to a difficulty level, which corresponds to the task's sequence length, of $10,000$, and evaluate on longer sequences. The SAM models (with and without ANN) are able to perform much better than chance (48 bits) on sequences of length $200,000$.
    }
    \label{fig:extrapolation}
\end{figure}

\section{Babi results}
\label{sec:babi}

See the main text for a description of the Babi task and its relevance. Here we report the best and mean results for all of the models on this task.

The task was encoded using straightforward 1-hot word encodings for both the input and output. We trained a single model on all of the tasks, and used the 10,000 examples per task version of the training set (a small subset of which we used as a validation set for selecting the best run and hyperparameters). Previous work has reported best results (Supplementary table \ref{tab:babi:best}), which with only 15 runs is a noisy comparison, so we additionally report the mean and variance for all runs with the best selected hyperparameters (Supplementary table \ref{tab:babi:mean}).

\begin{landscape}

\begin{table}[]
    \centering
    \begin{tabular}{l|c|c|c|c|c|c|c|c}

 & LSTM & DNC & SDNC & DAM & SAM & NTM & MN S & MN U \\
\hline
 1:    1 supporting fact  & $ 28.8$ & $  0.0$ & $  0.0$ & $  0.0$ & $  0.0$ & $ 16.4$ & $  0.0$ & $  0.0$ \\
 2:   2 supporting facts  & $ 57.3$ & $  3.2$ & $  0.6$ & $  0.2$ & $  0.2$ & $ 56.3$ & $  0.0$ & $  1.0$ \\
 3:   3 supporting facts  & $ 53.7$ & $  9.5$ & $  0.7$ & $  1.3$ & $  0.5$ & $ 49.0$ & $  0.0$ & $  6.8$ \\
 4: 2 argument relations  & $  0.7$ & $  0.0$ & $  0.0$ & $  0.0$ & $  0.0$ & $  0.0$ & $  0.0$ & $  0.0$ \\
 5: 3 argument relations  & $  3.5$ & $  1.7$ & $  0.3$ & $  0.4$ & $  0.7$ & $  2.5$ & $  0.3$ & $  6.1$ \\
 6:     yes/no questions  & $ 17.6$ & $  0.0$ & $  0.0$ & $  0.0$ & $  0.0$ & $  9.6$ & $  0.0$ & $  0.1$ \\
 7:             counting  & $ 18.5$ & $  5.3$ & $  0.2$ & $  0.4$ & $  1.9$ & $ 12.0$ & $  3.3$ & $  6.6$ \\
 8:           lists/sets  & $ 20.9$ & $  2.0$ & $  0.2$ & $  0.0$ & $  0.4$ & $  6.5$ & $  1.0$ & $  2.7$ \\
 9:      simple negation  & $ 18.2$ & $  0.1$ & $  0.0$ & $  0.0$ & $  0.1$ & $  7.0$ & $  0.0$ & $  0.0$ \\
10: indefinite knowledge  & $ 34.0$ & $  0.6$ & $  0.2$ & $  0.0$ & $  0.2$ & $  7.6$ & $  0.0$ & $  0.5$ \\
11:    basic coreference  & $  9.0$ & $  0.0$ & $  0.0$ & $  0.0$ & $  0.0$ & $  2.5$ & $  0.0$ & $  0.0$ \\
12:          conjunction  & $  5.5$ & $  0.1$ & $  0.1$ & $  0.0$ & $  0.1$ & $  4.6$ & $  0.0$ & $  0.1$ \\
13: compound coreference  & $  6.3$ & $  0.4$ & $  0.1$ & $  0.0$ & $  0.0$ & $  2.0$ & $  0.0$ & $  0.0$ \\
14:       time reasoning  & $ 56.1$ & $  0.2$ & $  0.1$ & $  3.8$ & $  4.3$ & $ 44.2$ & $  0.0$ & $  0.0$ \\
15:      basic deduction  & $ 49.3$ & $  0.1$ & $  0.0$ & $  0.0$ & $  0.0$ & $ 25.4$ & $  0.0$ & $  0.2$ \\
16:      basic induction  & $ 53.2$ & $ 51.9$ & $ 54.1$ & $ 52.8$ & $ 53.1$ & $ 52.2$ & $  0.0$ & $  0.2$ \\
17: positional reasoning  & $ 41.7$ & $ 21.7$ & $  0.3$ & $  6.0$ & $ 16.0$ & $ 39.7$ & $  0.0$ & $ 41.8$ \\
18:       size reasoning  & $  8.4$ & $  1.8$ & $  0.1$ & $  0.3$ & $  1.1$ & $  3.6$ & $ 24.6$ & $  8.0$ \\
19:         path finding  & $ 76.4$ & $  4.3$ & $  1.2$ & $  1.5$ & $  2.6$ & $  5.8$ & $  2.1$ & $ 75.7$ \\
20:  agent's motivations  & $  1.9$ & $  0.1$ & $  0.0$ & $  0.1$ & $  0.0$ & $  2.2$ & $ 31.9$ & $  0.0$ \\
\hline
Mean Error (\%)  & $ 28.0$ & $  5.2$ & $  2.9$ & $  3.3$ & $  4.1$ & $ 17.5$ & $  3.2$ & $  7.5$ \\
Failed tasks (err. > 5\%) & $   17$ & $    4$ & $    1$ & $    2$ & $    2$ & $   13$ & $    2$ & $    6$ \\
    \end{tabular}
    \caption{Test results for the best run (chosen by validation set) on the Babi task. The model was trained and tested jointly on all tasks. All tasks received approximately equal training resources. Both SAM and DAM pass all but 2 of the tasks, without any supervision of their memory accesses. SDNC achieves the best reported result on this task with unsupervised memory access, solving all but 1 task.
    We've included comparison with memory networks, both with supervision of memories (MemNet S) and, more directly comparable with our approach, learning end-to-end (MemNets U). }
    \label{tab:babi:best}
\end{table}

\begin{table}[h]
    \centering
    \begin{tabular}{l|c|c|c|c|c|c}
 & LSTM & DNC & SDNC & DAM & SAM & NTM \\
\hline
 1:    1 supporting fact  & $ 30.9 \pm   1.5$ & $  2.2 \pm   5.6$ & $  0.0 \pm   0.0$ & $  2.9 \pm  10.7$ & $  4.7 \pm  12.8$ & $ 31.5 \pm  15.3$ \\
 2:   2 supporting facts  & $ 57.4 \pm   1.2$ & $ 23.9 \pm  21.0$ & $  7.1 \pm  14.6$ & $ 12.1 \pm  19.3$ & $ 30.9 \pm  25.1$ & $ 57.0 \pm   1.3$ \\
 3:   3 supporting facts  & $ 53.0 \pm   1.4$ & $ 29.7 \pm  15.8$ & $  9.4 \pm  16.7$ & $ 15.3 \pm  17.4$ & $ 31.4 \pm  21.6$ & $ 49.4 \pm   1.3$ \\
 4: 2 argument relations  & $  0.7 \pm   0.4$ & $  0.1 \pm   0.1$ & $  0.1 \pm   0.1$ & $  0.1 \pm   0.1$ & $  0.2 \pm   0.2$ & $  0.4 \pm   0.3$ \\
 5: 3 argument relations  & $  4.9 \pm   0.9$ & $  1.3 \pm   0.3$ & $  0.9 \pm   0.3$ & $  1.0 \pm   0.4$ & $  1.0 \pm   0.5$ & $  2.7 \pm   1.2$ \\
 6:     yes/no questions  & $ 18.8 \pm   1.0$ & $  2.8 \pm   5.0$ & $  0.1 \pm   0.2$ & $  1.9 \pm   5.3$ & $  3.9 \pm   6.7$ & $ 18.6 \pm   2.7$ \\
 7:             counting  & $ 18.2 \pm   1.1$ & $  7.3 \pm   5.9$ & $  1.6 \pm   0.9$ & $  4.5 \pm   6.1$ & $  7.3 \pm   6.6$ & $ 18.7 \pm   3.2$ \\
 8:           lists/sets  & $ 20.9 \pm   1.4$ & $  4.0 \pm   4.1$ & $  0.5 \pm   0.4$ & $  2.7 \pm   5.4$ & $  3.6 \pm   6.2$ & $ 18.5 \pm   5.9$ \\
 9:      simple negation  & $ 19.4 \pm   1.5$ & $  3.0 \pm   5.2$ & $  0.0 \pm   0.1$ & $  2.1 \pm   5.5$ & $  3.8 \pm   6.7$ & $ 17.6 \pm   3.4$ \\
10: indefinite knowledge  & $ 33.0 \pm   1.6$ & $  3.2 \pm   5.9$ & $  0.3 \pm   0.2$ & $  3.4 \pm   8.1$ & $  5.7 \pm   9.2$ & $ 25.6 \pm   6.9$ \\
11:    basic coreference  & $ 15.9 \pm   3.3$ & $  0.9 \pm   3.0$ & $  0.0 \pm   0.0$ & $  1.5 \pm   5.5$ & $  2.6 \pm   7.9$ & $ 15.2 \pm   9.4$ \\
12:          conjunction  & $  7.0 \pm   1.3$ & $  1.5 \pm   1.6$ & $  0.2 \pm   0.3$ & $  1.8 \pm   6.4$ & $  2.9 \pm   7.9$ & $ 14.7 \pm   8.9$ \\
13: compound coreference  & $  9.1 \pm   1.4$ & $  1.5 \pm   2.5$ & $  0.1 \pm   0.1$ & $  0.6 \pm   2.2$ & $  1.3 \pm   2.4$ & $  6.8 \pm   3.3$ \\
14:       time reasoning  & $ 57.0 \pm   1.6$ & $ 10.6 \pm   9.4$ & $  5.6 \pm   2.9$ & $ 11.5 \pm  15.0$ & $ 15.0 \pm  12.6$ & $ 52.6 \pm   5.1$ \\
15:      basic deduction  & $ 48.1 \pm   1.3$ & $ 31.3 \pm  15.6$ & $  3.6 \pm  10.3$ & $ 17.2 \pm  19.7$ & $  5.5 \pm  13.8$ & $ 42.0 \pm   6.9$ \\
16:      basic induction  & $ 53.8 \pm   1.4$ & $ 54.0 \pm   1.9$ & $ 53.0 \pm   1.3$ & $ 53.8 \pm   1.0$ & $ 53.6 \pm   1.2$ & $ 53.8 \pm   2.1$ \\
17: positional reasoning  & $ 40.8 \pm   1.8$ & $ 27.7 \pm   9.4$ & $ 12.4 \pm   5.9$ & $ 16.9 \pm  10.3$ & $ 20.4 \pm   8.6$ & $ 40.1 \pm   1.3$ \\
18:       size reasoning  & $  7.3 \pm   1.9$ & $  3.5 \pm   1.5$ & $  1.6 \pm   1.1$ & $  1.8 \pm   1.7$ & $  3.0 \pm   1.8$ & $  5.0 \pm   1.2$ \\
19:         path finding  & $ 74.4 \pm   1.3$ & $ 44.9 \pm  29.0$ & $ 30.8 \pm  24.2$ & $ 23.0 \pm  25.4$ & $ 33.7 \pm  27.8$ & $ 60.8 \pm  24.6$ \\
20:  agent's motivations  & $  1.7 \pm   0.4$ & $  0.1 \pm   0.2$ & $  0.0 \pm   0.0$ & $  0.1 \pm   0.5$ & $  0.0 \pm   0.0$ & $  2.0 \pm   0.3$ \\
\hline
Mean Error (\%)  &
$ 28.7 \pm  0.5$ & $ 12.8 \pm  4.7$ & $  6.4 \pm  2.5$ & $  8.7 \pm  6.4$ & $ 11.5 \pm  5.9$ & $ 26.6 \pm  3.7$ \\
Failed tasks (err. > 5\%) & $ 17.1 \pm  0.8$ & $  8.2 \pm  2.5$ & $  4.1 \pm  1.6$ & $  5.4 \pm  3.4$ & $  7.1 \pm  3.4$ & $ 15.5 \pm  1.7$ \\
    \end{tabular}
    \caption{Mean and variance of test errors for the best set of hyperparameters (chosen according the validation set). Statistics are generated from 15 runs.}
    \label{tab:babi:mean}
\end{table}

\end{landscape}

\end{document}